%% file: neurips_data_2022.tex
\documentclass{article}
\PassOptionsToPackage{numbers, compress}{natbib}

\usepackage[final]{neurips_data_2022}
\usepackage [table]{xcolor}
\usepackage[utf8]{inputenc} 
\usepackage[T1]{fontenc}    
\usepackage{hyperref}       
\usepackage{url}            
\usepackage{booktabs}       
\usepackage{amsfonts}       
\usepackage{nicefrac}       
\usepackage{microtype}      
\usepackage{xcolor}         

\definecolor{lightblue}{RGB}{72, 187, 231}
\newcolumntype{a}{>{\columncolor{lightblue!15}}c}

\usepackage{graphicx, rotating}
\usepackage{booktabs}
\usepackage{multirow}
\usepackage{comment}
\usepackage{amsmath}
\usepackage{xspace}
\usepackage{hyperref}
\definecolor{darkblue}{rgb}{0,0.3,0.99}
\newcommand{\vae}[1]{{\color{magenta}#1}}

\title{3DOS: Towards 3D Open Set Learning -- \\Benchmarking and Understanding\\ Semantic Novelty Detection on Point Clouds}

\author{
  Antonio Alliegro\thanks{Equal contribution} , Francesco Cappio Borlino\footnotemark[1] , Tatiana Tommasi\\
  Department of Control and Computer Engineering. 
  Politecnico di Torino, Italy\\
  Italian Institute of Technology, Italy \\
  \texttt{\{antonio.alliegro, francesco.cappio, tatiana.tommasi\}@polito.it}\\
}

\usepackage{amsmath,amssymb}

\input{symbols}
\begin{document}
\maketitle

\begin{abstract}
\input{sections/abstract}
\end{abstract}
\section{Introduction}
\input{sections/intro}
\section{Related Work}
\label{sec:related}
\input{sections/related}
\section{3DOS Benchmark}
\input{sections/benchmarkd}
\section{Experiments}
\input{sections/experiments}

\section{Conclusions}
\input{sections/conclusion}
\bibliographystyle{abbrv}
\bibliography{egbib}
\newpage
\appendix
\input{sections_suppl/supp-details}
\input{sections_suppl/supp-synth-real}
\input{sections_suppl/supp-error_bars}

\input{sections_suppl/supp-limit}
\end{document}

%% file: symbols.tex
\newcommand{\bx}{\boldsymbol{x}}

\newcommand{\sY}{\mathcal{Y}}

\newcommand{\sS}{\mathcal{S}}
\newcommand{\sT}{\mathcal{T}}

\newcommand{\eucli}{CE $(L^2)$\xspace}
\newcommand{\rw}{real-world\xspace}

\makeatletter

\usepackage{accents}

\usepackage{xcolor}
\usepackage{multirow}

\makeatother

%% file: sections/abstract.tex
In recent years there has been significant progress in the field of 3D learning on classification, detection and segmentation problems. The vast majority of the existing studies focus on canonical closed-set conditions, neglecting the intrinsic open nature of the \rw. 
This limits the abilities of robots and autonomous systems involved in safety-critical applications that require managing novel and unknown signals.
In this context exploiting 3D data can be a valuable asset since it provides rich information about the geometry of perceived objects and scenes.
With this paper we provide the first broad study on 3D Open Set learning. 
We introduce 3DOS: a novel testbed for semantic novelty detection that considers several settings with increasing difficulties in terms of semantic (category) shift, and covers both in-domain (synthetic-to-synthetic, real-to-real) and cross-domain (synthetic-to-real) scenarios. 
Moreover, we investigate the related 2D Open Set literature to understand if and how its recent improvements are effective on 3D data. Our extensive benchmark positions several algorithms in the same coherent picture, revealing their strengths and limitations. 
The results of our analysis may serve as a reliable foothold for future tailored 3D Open Set methods.

%% file: sections/intro.tex
\label{sec:intro}
Most existing machine learning models rely on the assumption that train and test data are drawn \textit{i.i.d.} from the same distribution. While reasonable for lab experiments, this assumption frequently fails to hold when models are deployed in the open world, where a variety of distributional shifts with respect to the training data can emerge. For example, new object categories may induce a semantic shift, or data from new domains may give rise to a covariate shift \cite{yang2021scood,yang2021oodsurvey,ruan2022optimal}. Such cases can occur separately or jointly, and the test samples that differ from what was observed during training are generally indicated as out-of-distribution (OOD) data.
These data may become extremely dangerous for autonomous agents as testified by the numerous accidents involving self-driving cars that misbehaved 
when encountering anomalous objects in the streets\footnote{https://edition.cnn.com/2021/08/27/cars/toyota-self-driving-vehicle-paralympics-accident/index.html https://www.foxnews.com/auto/tesla-smashes-overturned-truck-autopilot}. 
To avoid similar risks it is of paramount importance to build robust models capable of maintaining their discrimination ability over the closed set of known classes while rejecting unknown categories. Solving this task is challenging for existing deep models: their exceptional closed set performance hides miscalibration \cite{calibration_deep_nets} and over-confidence issues \cite{nguyen2015deep}. In other words, their output score cannot be regarded as a reliable measure of prediction correctness. 
This drawback has been largely discussed in the 2D visual learning literature 
\cite{react,ARPL_TPAMI, OpenHybrid, vaze2022openset} as its solution would enable the use of powerful deep models for many \rw tasks. 
In this context, and particularly for many safety-critical applications such as self-driving cars, 3D sensing is a valuable asset, providing detailed information about the geometry of sensed objects that 2D images cannot capture. 
However 3D literature in this field is still in its infancy, with only a small number of works which have just started to scratch the surface of the problem by focusing on particular sub-settings \cite{towards_japan, biplab1}. 
With this work, we draw the community's attention to 3D Open Set learning, which entails developing models designed to process 3D point clouds that can recognize test samples from a set of known categories while avoiding prediction for samples from unknown classes. 
Our contributions are: 
1) we propose 3DOS, the first benchmark for 3D Open Set learning, considering several settings with increasing levels of difficulty. 
It includes three main tracks: Synthetic, Real to Real, and Synthetic to Real. The first is meant to investigate the behavior of existing Open Set methods on 3D data, the other two are designed to simulate \rw deployment conditions; 
2) we build a coherent picture by putting together the existing literature from OOD detection and Open Set recognition in 2D and 3D;
3) we analyze the performance of these methods to discover which is the state-of-the-art for 3D Open Set learning. We highlight their advantages and limitations and show that often a simple representation learning approach is enough to outperform sophisticated state-of-the-art methods.\\
Our code and data are available at \url{https://github.com/antoalli/3D_OS}. 

%% file: sections/related.tex
\begin{figure}[t]
    \centering
    \includegraphics[width=1\textwidth]{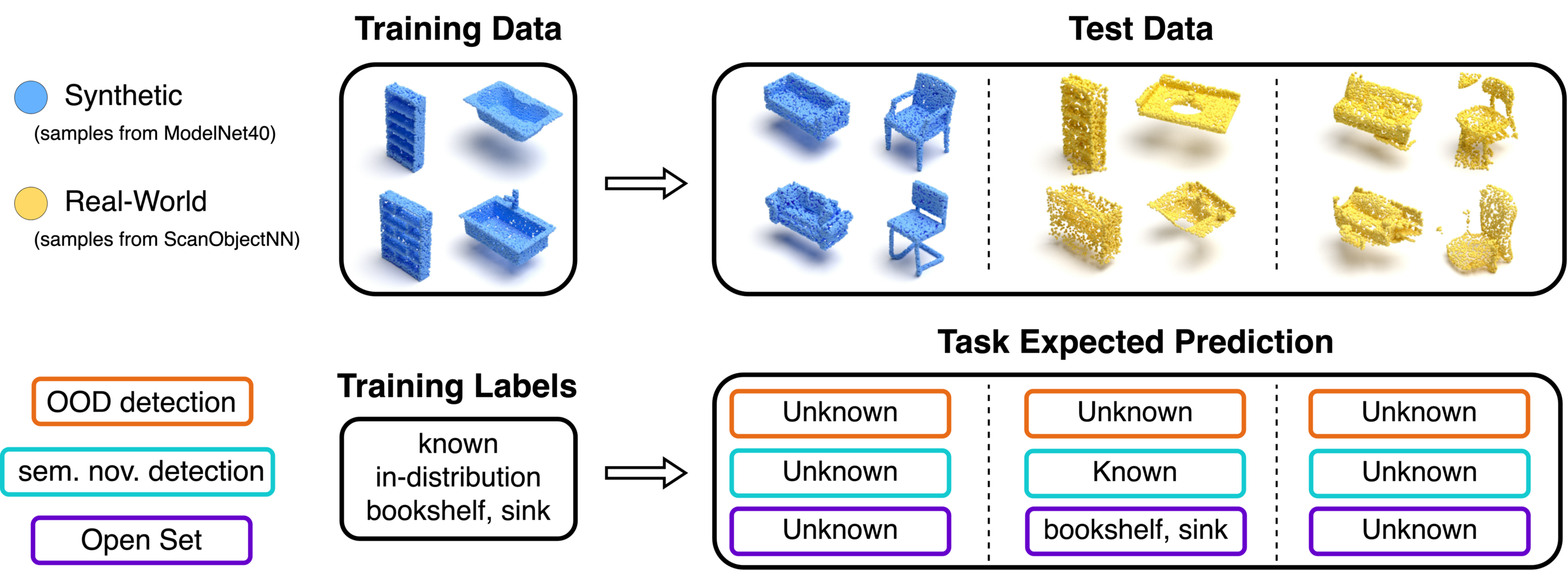}
    \caption{Schematic illustration of the \textit{OOD detection}, \textit{semantic novelty detection} and \textit{Open Set} tasks on 3D data.\label{fig:settings}}
    \vspace{-3mm}
\end{figure}

We provide an overview of existing literature on \emph{OOD detection} and \emph{Open Set learning}. The difference between these two tasks is often neglected, but it is important to point it out (see Fig. \ref{fig:settings}). 
In OOD detection it is sufficient to identify and reject samples with any distribution shift with respect to the training data. 
In the particular case of \emph{semantic novelty detection}, the concept of novelty is limited to the categories not seen during training, regardless of  the specific domain appearance of the observed samples. Besides separating data of known classes from those of unknown classes, Open Set recognition requires performing a class prediction over the known categories.

\textbf{Discriminative Methods.}
By training a model with multi-class supervision we expect to get low uncertainty on in-distribution (ID) data and high uncertainty for OOD samples. Thus, a baseline approach may consist in using the maximum softmax probability (MSP) as a \emph{normality score} to separate known and unknown instances \cite{MSP}. However, deep models suffer from over-confidence \cite{nguyen2015deep} and their prediction outputs need some re-calibration to be considered as uncertainty scoring functions. ODIN \cite{ODIN} exploited temperature scaling and input pre-processing to better separate ID from OOD samples. In \cite{energy} the authors showed how to derive Energy scores from the prediction output, demonstrating that they are better aligned with the probability density of the inputs and are less prone to over-confidence. Instead of considering the output, GradNorm \cite{gradnorm} focused on the network's gradients showing that their norm carries distinctive signatures to
amplify the ID/OOD separability. 
ReAct \cite{react} proposed to further increase this separability by rectifying the internal network activations.
Finally, a very recent work has discussed how the normalized softmax probabilities can be replaced by the
maximum logit scores (MLS), resulting in an approach competitive with other more complex strategies  \cite{vaze2022openset}.
We highlight that all the methods of this discriminative family are applied post-hoc on the closed set classifier, meaning that the original training procedure and objective are not modified. Thus, the models
maintain their ability to distinguish among the known classes and are suitable for Open Set recognition. 

\textbf{Density and Reconstruction Based Methods.} 
Density-based methods are trained to model the distribution of known data~\cite{GPointNet,OpenHybrid}. Input samples are then identified as unknown if lying in low-likelihood regions. Several works have exploited generative models for OOD detection with novelty metrics that range from basic sample reconstruction \cite{abati2019latent,Choi2020Novelty} to more complex likelihood ratio and regret \cite{NEURIPS2019_ratios,NEURIPS2020_eddea82a}. Still, generative models can be difficult to train, and their performance is frequently lower than that of discriminative ones. Recently a hybrid approach proposed to combine discriminative and probabilistic flow-based learning with promising results \cite{OpenHybrid}.

\textbf{Outlier Exposure.}
Another line of OOD approaches exploits \emph{outlier} data available at training time. They are used to regularize the model by applying conditions on the prediction entropy \cite{outlier_exposure,MCD} or running outlier mining, re-sampling and filtering  \cite{chen2021atom,Li_2020_CVPR,yang2021scood}.

\textbf{OOD Data Generation.}
In many practical cases, it is not possible to access outlier samples at training time. Thus, unknown sample synthesis is used to prepare the model for the deployment conditions  \cite{counterfactual,du2022vos,ARPL_TPAMI,OE_mixup}. Some recent OOD approaches have also combined real outlier mining and fake outlier generation \cite{Kong_2021_ICCV}.

\textbf{Representation and Distance Based Methods}.
Enhancing data representation may help to better characterize known data and consequently ease the identification of unknown samples. In a reliable embedding space, OOD samples should be far away from ID classes so that the distance from stored exemplars or prototypes can be used as a scoring function. 
Existing approaches focus on two aspects: how to learn a good representation and how to measure distances. Self-supervised, contrastive and prototypes learning methods are of the first kind and generally rely on cosine similarity \cite{tack2020_CSI,DRP_eccv2020,podn_scientificreport}. Other solutions build on discriminative models, but instead of considering the prediction output, they focus on the learned features and evaluate sample distances by using different metrics like $L^2$ norm, layer-wise Mahalanobis, or similarity metrics based on Gram matrices \cite{huang2021Euclideanfeature,NEURIPS2018_abdeb6f5,gramICML2020}. 

All the references mentioned above come from the 2D literature. Up to our knowledge 3D OOD detection and Open Set problems have been studied only by a handful of works. 
A VAE approach for reconstruction-based 3D OOD detection is provided in \cite{towards_japan}, together with an analysis on seven classes of the ShapeNet dataset \cite{shapenet_dataset}, each used in turn as unknown. The study considers different VAE normality scores but does not compare with other baselines.
In \cite{biplab1} the authors distilled knowledge from a large teacher network while also adding data produced by mixing training samples to define an unknown class. The authors focus on building a lightweight model and do not include comparisons with other Open Set methods. Moreover the classes in the used known/unknown datasets (ModelNet10/40 \cite{mn40_dataset}) significantly differ in pose and headings which makes their separation trivial \cite{Lee_2021_CVPR}. 
Two other works refer to 3D Open Set object detection and segmentation, but their objective is mainly clustering to aggregate points into object instances \cite{3dv-opensetdetection,WongWRLU19Urtasun}.

%% file: sections/benchmarkd.tex
\begin{figure*}[t!]
    \centering
    \includegraphics[width=1\textwidth]{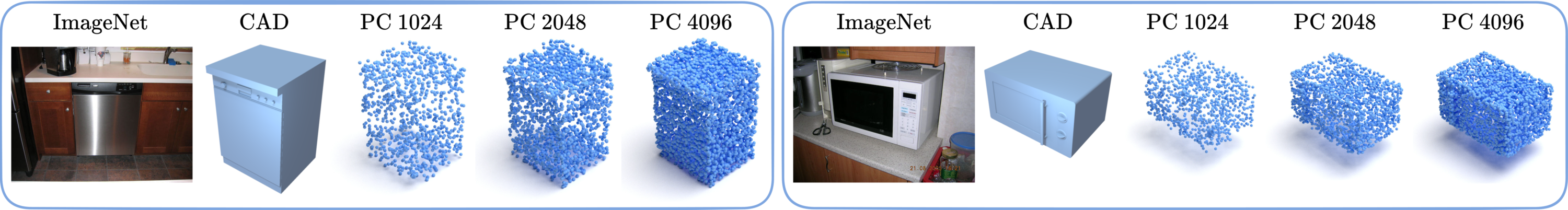}
    \caption{By looking at the point clouds of a dishwasher and microwave it might be very difficult to understand if they are the same object or not. Differently from the images, point clouds capture the object geometry, but they miss the original scale as well as color, texture, and object context which are naturally present in images.\label{fig:2Dvs3D}}
    \vspace{-4mm}
\end{figure*}

Object recognition on 3D data is much more challenging 
than on 2D samples, with the main issues originating from the lack of color, texture, and of the general context in which the objects usually appear in images (see Fig. \ref{fig:2Dvs3D}). When the goal is to evaluate whether a certain instance belongs to a  known or novel class, all these cues provide crucial pieces of evidence, but without them, the task becomes really difficult.
Data rescaling, resolution and noise can also significantly influence the final prediction. 
We dedicate our work to investigating all these aspects within the task of 3D Open Set learning: in the following we formalize the problem, introduce several testbeds and present an extensive experimental analysis.

\subsection{Preliminaries}
\textbf{Problem formulation.}
We consider the labeled set $\sS=\{\bx^{s},y^{s}\}_{s=1}^{N}$ drawn from the training distribution $p_{\sS}$, and we indicate as \emph{known} all the classes $y^{s}\in\sY^s$ covered by this set.  A model trained on $\sS$ is later evaluated on the test set $\sT = \{\bx^{t}\}_{t=1}^{M}$ drawn from the distribution $p_{\sT}$.  In the Open Set scenario train and test distributions differ in terms of semantic content: for the test data labels $y^{t}\in\sY^t$ it holds $\sY^s \neq \sY^t$. More specifically, we consider a partial overlap between the two sets $\sY^s \subset \sY^t$ and the test classes which do not appear in the \textit{known} classes set $\sY^s$ are therefore \textit{unknown}. 
A reliable semantic novelty detection model trained on $\sS$ should output for each test sample $\bx^t$ a normality score representing its probability of belonging to any of the known classes in $\sY^s$. An Open Set model must also provide an output probability distribution over each of the classes in $\sY^s$.

\textbf{Performance Metrics.} 
We evaluate the ability to detect unknown samples in test data by exploiting two metrics: \textbf{AUROC} and \textbf{FPR95}. 
Given that the detection of unknown samples is a binary task, both metrics are based on the concepts of True Positive (TP), False Positive (FP), True Negative (TN), and False Negative (FN). The AUROC (the higher the better) is the Area Under the Receiver Operating Characteristic Curve. The ROC curve is a graph showing the TP rate (TPR) and the FP rate (FPR) plotted against each other \cite{MSP} when varying the normality threshold. As a result, the AUROC is a threshold-free metric, and it can be interpreted as the probability that a known test sample has a greater normality score than an unknown one. 
The FPR95 (the lower the better) is the FP Rate at TP Rate 95\%, sometimes referred to as FPR@TPRx with x=95\%. This metric is based on the choice of a normality threshold so that 95\% of positive samples are predicted as positives (TPR=TP/TP+FN). Then the false positive rate (FPR=FP/FP+TN) is computed using this threshold.
For Open Set methods we also evaluate their ability to correctly classify known data by computing their classification accuracy (\textbf{ACC}). Further metrics are reported in the supplementary material.

\textbf{Datasets.}
We build the 3DOS Benchmark on top of three well known 3D objects datasets:
ShapeNetCore~\cite{shapenet_dataset}, ModelNet40~\cite{mn40_dataset} and ScanObjectNN~\cite{scanobjectnn_ICCV19}. 

\emph{ShapeNetCore} contains 51,127 meshes from synthetic instances of 55 common object categories. 
In our analysis we adopt ShapeNetCore v2 and use the official training (70\%), validation (10\%) and test (20\%) splits. 
All objects are consistently aligned in pose. 
Having consistent alignment between different semantic categories is fundamental to avoid any bias that could lead to trivial inter-categories discrimination.
The point clouds are obtained by uniformly sampling  points from the mesh surface and normalized to fit within a unit cube centered at the origin. 
In our analysis we merge telephone and cellphone categories since they share similar semantic content, thus obtaining a total of 54 categories.\\
\emph{ModelNet40}~\cite{mn40_dataset} contains 12311 3D CAD models from 40 man-made object categories. We use the official dataset split, consisting of 9843 train and 2468 test shapes according to \cite{pointnet++}. We obtain a point cloud from each CAD model by uniformly sampling points from the faces of the synthetic mesh. Each point cloud is then centered in the origin and scaled to fit within a unit cube.\\
\emph{ScanObjectNN}~\cite{scanobjectnn_ICCV19} contains 2902 3D scans of \rw objects from 15 categories. Specifically, we consider the original \texttt{OBJ\_BG} split in which 3D scans are affected by acquisition artifacts such as vertex noise, non-uniform density, missing parts and occlusions. 
Data samples are already in the form of point clouds with 2048 points each and include the foreground object as well as background and other interacting objects which are absent in the synthetic instances of ModelNet and ShapeNet.

\subsection{Benchmark Tracks} 

3DOS includes three main Open Set tracks. The \emph{Synthetic Benchmark} is designed to assess the performance of existing methods in the presence of semantic shift, while the more challenging \emph{Synth to Real Benchmark} covers both semantic and domain shift, with train and test samples that are respectively drawn from synthetic data~(Modelnet40) and \rw data~(ScanObjectNN). Finally, the \emph{Real to Real Benchmark} represents an intermediate case with semantic shift among training and test data and noisy samples (from ScanObjectNN) in both sets.

\begin{figure*}[t!]
    \centering
    \includegraphics[width=1\textwidth]{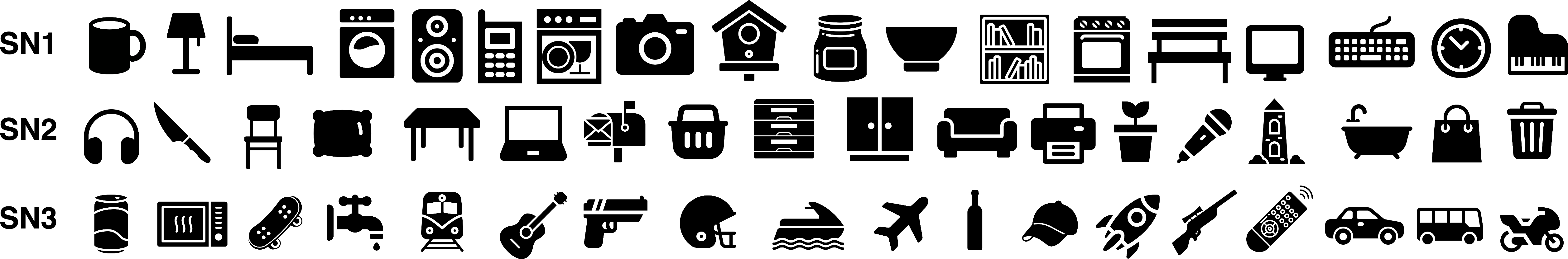}
    \caption{Visualization of the object categories in each of the sets of the Synthetic Benchmark. \textbf{SN1}: mug, lamp, bed, washer, loudspeaker, telephone, dishwasher, camera, birdhouse, jar, bowl, bookshelf, stove, bench, display, keyboard, clock, piano. \textbf{SN2}: earphone, knife, chair, pillow, table, laptop, mailbox, basket, file cabinet, sofa, printer, flowerpot, microphone, tower, bag, trash bin. \textbf{SN3}: can, microwave, skateboard, faucet, train, guitar, pistol, helmet, watercraft, airplane, bottle, cap, rocket, rifle, remote, car, bus, motorbike.\label{fig:icons_synth}}
    \vspace{-2mm}
\end{figure*}

\begin{figure*}[t!]
    \centering
    \includegraphics[width=1\textwidth]{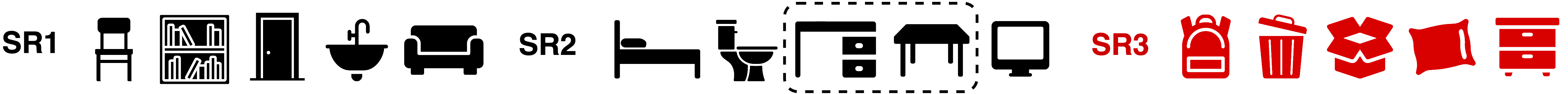}
    \caption{Visualization of the object categories in each of the sets of the Synthetic to Real Benchmark. \textbf{SR1}: chair, shelf, door, sink, sofa. \textbf{SR2}: bed, toilet, desk, table, display. \textbf{SR3}: bag, bin, box, pillow, cabinet.\label{fig:icons_synth2real}}
    \vspace{-4mm}
\end{figure*}

\textbf{Synthetic Benchmark.} For our synthetic testbed we employ ShapeNetCore~\cite{shapenet_dataset} dataset and we split it into 3 not overlapping (\emph{i.e.} semantically different) category sets of 18 categories each. We dub them as SN1, SN2 and SN3 (see Fig. \ref{fig:icons_synth} for the list of categories belonging to each set).\\

We obtained three scenarios of increasing difficulty by simply selecting each of the SN-Sets in turn as \textit{known} class set and considering the remaining two category sets as \textit{unknown}.
For this track models are trained on the train split of the known classes set and evaluated on the test split of both known and unknown classes.

\textbf{Synthetic to Real Benchmark.} To define our Synthetic to Real-World cross-domain scenario, we employ synthetic point clouds from ModelNet40~\cite{mn40_dataset} for training while we test on \rw point clouds from ScanObjectNN~\cite{scanobjectnn_ICCV19}.
We choose to adopt ModelNet40 (instead of ShapeNetCore) because it has a better overlap with ScanObjectNN and previous works already considered the same cross-domain scenario in the context of point cloud object classification \cite{alliegro_icpr, cardace2021refrec}.
We define three different category sets: SR1, SR2, and SR3 as described in Fig. \ref{fig:icons_synth2real}.
The first two sets are composed by matching classes of ModelNet40 and ScanObjectNN. The third set (SR3) is instead composed by ScanObjectNN classes without such a one-to-one mapping with ModelNet40. Overall we have two scenarios with either SR1 or SR2 used as \textit{known} and the other two considered as \emph{unknown}.
For this track, models are trained on ModelNet40 samples of the known classes  set and evaluated on the ScanObjectNN samples of both known and unknown classes.

\textbf{Real to Real Benchmark.} For this last case we exploited the same SR category sets created from ScanObjectNN described above. Specifically, each of them is used as \textit{unknown} in the test set, while the other two are divided into train and test and used as \textit{known} classes.

\subsection{Evaluated Methods}
\label{sec:benchmark_methods}
We consider several approaches from the families of methods described in Sec. \ref{sec:related}.

\textbf{Discriminative Methods.} 
All these methods are built on top of a standard closed set classifier trained with cross-entropy.
For our analysis we select the \textbf{MSP}~\cite{MSP} baseline, as well as its maximum logit score variant (\textbf{MLS})~\cite{vaze2022openset}. We further consider \textbf{ODIN}~\cite{ODIN}, \textbf{Energy}~\cite{energy}, \textbf{GradNorm}~\cite{gradnorm} and \textbf{ReAct}~\cite{react}. 

\textbf{Density and Reconstruction Based Methods.} We select two methods from this group. We test a \textbf{VAE} model with reconstruction based scoring by following one of the few existing works on 3D anomaly detection \cite{towards_japan}. This is the only unsupervised approach in our analysis, and thus performs only OOD detection without providing predictions over known classes. 
The second approach is based on Normalizing Flow (\textbf{NF}).
We took inspiration by the 2D Open Set state-of-the art OpenHybrid~\cite{OpenHybrid} and the anomaly detection method DifferNet~\cite{RudWan2021differnet}. Specifically we train a NF model consisting of 8 coupling blocks~\cite{realnvp} on top of the same feature embedding used by a cross-entropy classifier. The training objective consists in the maximization of the log-likelihood of training samples, and the predicted log-likelihood is later used to distinguish ID and OOD samples. Differently from the VAE this model also includes a closed set classifier and thus it is applicable to the Open Set task. 

\textbf{Outlier Exposure with OOD Generated Data.} Our analysis focuses on the setting where training data does not include unknown samples, thus we assess the performance of the OE approach presented in \cite{outlier_exposure} by exploiting fake OOD data produced via point cloud mixup~\cite{Lee_2021_CVPR} (\textbf{OE+mixup}). 

\textbf{Representation and Distance Based Methods.} To evaluate the effect of a carefully learned feature embedding on the identification of novel categories, we consider the state-of-the-art 2D Open Set method \textbf{ARPL+CS}~\cite{ARPL_TPAMI}. 
It learns reciprocal points that represent the \emph{otherness} with respect to each known class: the distance from these points is considered proportional to the probability that a sample belongs to a certain class. Moreover, the method includes confusing samples (CS) generated in an adversarial manner to represent samples of unseen classes which are equidistant from all the reciprocal points.
We also test the embedding of a cosine classifier (\textbf{Cosine proto}) as done in \cite{fontanel2021detecting}, this method learns class-prototypes by maximizing the cosine similarity between each training sample and the prototype of its class. At inference time the highest cosine similarity with a known class prototype is used as a normality score.
In order to learn features representation on closed set data it is also possible to use standard losses like the supervised cross-entropy or supervised contrastive~\cite{khosla2020supervised}. In the first case we rely on the euclidean distance between the feature of the test sample and the training samples~(\textbf{\eucli}), while in the second (\textbf{SupCon}) we use the cosine distance which better reflects the contrastive training objective.
In this analysis of distance based methods we also include a seemingly unconnected technique originally proposed for face recognition. The \textbf{SubArcFace} \cite{subarcface} approach belongs to the family of margin-based softmax methods that aim at simultaneously achieving maximal intra-class compactness and inter-class discrepancy without the drawbacks (negative sampling and large data batches) that affect the triplet and the contrastive losses.
With respect to other similar strategies \cite{arcface,cosface},
for each known class, SubArcFace identifies multiple sub-centers, and a training sample only needs to be close to one of them rather than to a single class prototype.
To get the normality score we follow the same procedure adopted for SupCon.
It should be noted that the last three methods listed (\eucli, SupCon, SubArcface) require training data to be available also at test time since they compute the test sample normality score as distance to the nearest train sample.

%% file: sections/experiments.tex
\input{tables/tables_synth}

We perform our experimental analysis with the goal of answering a set of research questions, each discussed below in separate paragraphs respectively for the 
Synthetic and Synthetic to Real benchmarks.
Given that 3D point clouds literature counts a large number of backbones with no dominant one, we perform all main experiments with two reliable backbones: DGCNN \cite{dgcnn} and PointNet++ \cite{pointnet++}.

\subsection{Implementation details}
\label{subsec:impl_details}
Unless otherwise specified, we use 1024 points for Synthetic point clouds (ShapeNet and ModelNet) and 2048 points for real-world point clouds (ScanObject).
In the Synthetic Benchmark we augment training data with scale and translation transformations, while for the Synthetic to Real case we also augment it through random rotation around the up-axis. 
All models are trained with a batch size of 64 for 250 epochs, with exception of SubArcFace which is trained for 500 epochs on synthetic sets (SN1, SN2, SN3).
For DGCNN experiments we use SGD optimizer with an initial learning rate of 0.1, momentum and weight decay are set respectively to 0.9 and 0.0001.
With PointNet++ we employ Adam optimizer and set the initial learning rate to 0.001.
Each experiment is repeated with three different seeds, we take results from the last epoch model and average across runs. Our code is implemented in PyTorch 1.9, experiments run on an HPC cluster with NVIDIA V100 GPUs.
All models are trained on a single GPU except for SupCon and ARPL methods. To facilitate reproducibility we provide a complete list and discussion of the analyzed methods hyperparameters in the supplementary material. 

\subsection{Synthetic Benchmark}
\textbf{How do OOD detection methods perform on 3D semantic novelty detection?} 
In these experiments we analyse the performance of OOD detection and Open Set methods on the  Synthetic Benchmark.
We report results in Tab. \ref{tab:synth2synth},
following the same four-group organization of Sec.\ref{sec:benchmark_methods}.

\emph{Discriminative Methods.} 
We consider MSP as the main baseline, but we include also its variant MLS~\cite{vaze2022openset}. We can see that the latter is a strong baseline, as it is often on par or better than more complex state-of-the-art methods (\emph{e.g.} ODIN, Energy, GradNorm).
In general, all methods of this group manage to improve the MSP baseline results both in terms of AUROC and FPR95, with the only exception of GradNorm. The peculiarity of this approach is that it relies on gradients extracted at test time from the network layers
to compute the normality score. We hypothesize that the substantial difference between 2D (for which the model has been originally designed) and 3D network architectures may be responsible for the observed poor performance. 
ReAct consistently outperforms all the others with both the DGCNN and PointNet++ backbones.

\emph{Density and Reconstruction Based Methods.}
VAE results are far below the MSP baseline, this could be expected since it is the only unsupervised method in the table. It should be noticed that its encoder matches neither with DGCNN nor with PointNet++. It is composed of graph convolutional layers while the decoder is inspired to FoldingNet~\cite{foldingnet}. 
We still include VAE results in Tab. \ref{tab:synth2synth} and Tab. \ref{tab:synth2real}, regardless of its peculiarities, but we report its numbers with a \vae{different color}. 
On the other hand, NF is quite sensitive to backbone choice; it performs well when trained on top of the semantically rich embedding extracted by the DGCNN, but it underperforms when trained on the local features embedding of PointNet++.

\emph{Outlier Exposure with OOD Generated Data.}  Comparing with the MSP baseline we observe that for both backbones the OE finetuning produces a slight improvement in terms of FPR95 while the AUROC does not show gains.

\emph{Representation and Distance Based Methods.} 
The ARPL+CS \cite{ARPL_TPAMI} approach is the current state-of-the-art 2D Open Set method, however the results indicate that it does not work as well on 3D data and it is easily outperformed by the much simpler Cosine proto.
Interestingly, the simple \eucli method built on a standard cross entropy classifier obtains promising results for both backbones outperforming all the competing methods.
The same considerations done for ARPL+CS hold for SupCon, which has been already successfully used in the past for OOD detection in 2D \cite{tack2020_CSI,sehwag2021_SSD}. 
We believe this is due to the fact that synthetic data are very clean and lack the variability required to build a good contrastive embedding. 
A better result can be obtained through SubArcFace which builds a similar feature embedding but is less computationally expensive and converges more easily. 
Given its state-of-the-art performance, for the following analyses we will primarily focus \eucli, along with MSP and MLS as baselines.

\textbf{What is the effect of improving closed-set classification on 3D semantic novelty detection?} A recent work has put under the spotlight the correlation between the closed accuracy of discriminative methods with their open set recognition performance on images \cite{vaze2022openset}. To verify this trend on 3D data we run two sets of experiments. \\
A first analysis is done by exploiting a standard regularization technique as label smoothing \cite{LS}. 
Tab. \ref{tab:LS} shows that LS provides a small closed set accuracy improvement for both backbones as well as some improvement also on AUROC and FPR95 when using the DGCNN backbone. It overcomes the results of ReAct (AUROC:76.4, FPR95:74.6), but remains still worse than Cosine proto which has the best performance on this set.
With PointNet++ the advantage is evident only in FPR95 for MLS and MSP, but the open set performance decreases both in terms of AUROC and FPR95 for \eucli.    
A second evaluation is done by changing the network backbone. We experiment with a range of distinct architectures beyond the already considered DGCNN and PointNet++. Specifically, we tested CurveNet~\cite{curvenet}, GDANet~\cite{gdanet}, RSCNN~\cite{RSCNN}, pointMLP~\cite{pointMLP} and PCT~\cite{pct}. In particular the latter exploits Transformer blocks for point cloud learning and has recently achieved state-of-the-art performance for 3D object classification and segmentation. The results in Fig. \ref{fig:scatter_backbones} (left) show that for various 3D backbones the open set performance is not strictly linked to the closed set one. In particular, while RSCNN reaches one of the top closed set accuracy results, its MSP AUROC is the worst one.

\textbf{Is 3D semantic novelty detection affected by the point cloud density?}
\begin{figure*}[t!]
    \centering
    \includegraphics[width=1\textwidth]{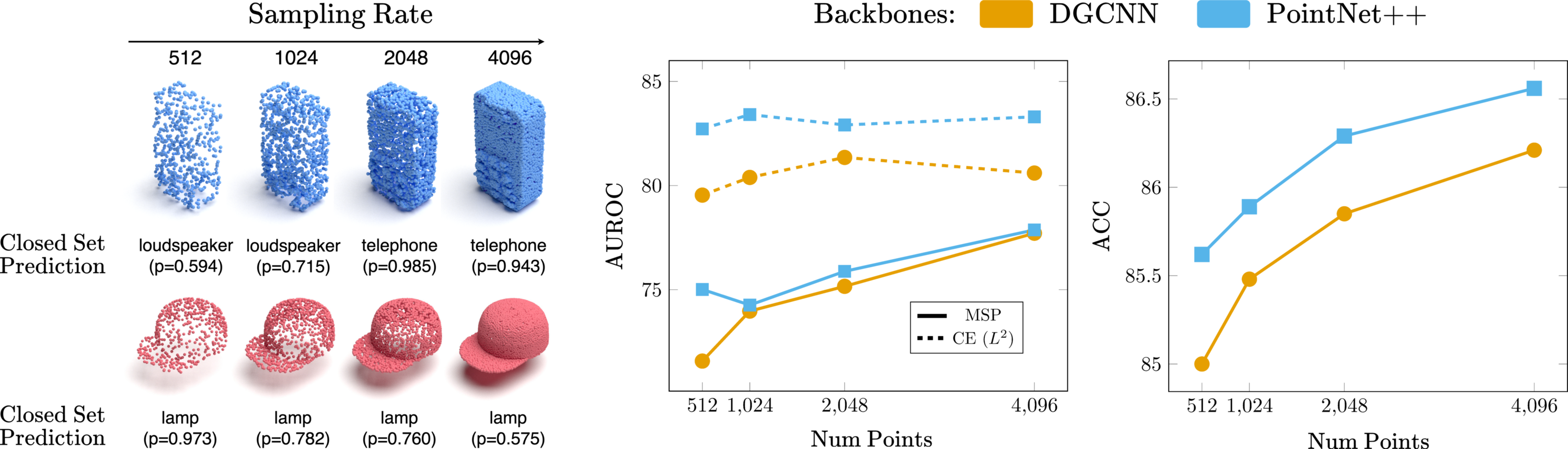}
    \caption{
    Analysis of the sampling rate influence on the hard SN1 set.
    Left: the blue telephone and red cap are known and unknown test samples for a DGCNN classifier trained on SN1. We show the class with the highest probability (p) assignment predicted by the classifier. The known object is correctly recognized as the sampling rate increases, while the confidence in the unknown object decreases, supporting rejection.
    Right: AUROC and ACC trends when varying the number of sampled points.\label{fig:pointsampling}}
    \vspace{-6mm}
\end{figure*}
We investigate the impact of the point cloud sampling rate.
We run experiments with different point cloud sizes: 512, 1024, 2048, and 4096. 
For each experiment we fix the number of points (\emph{e.g.} 512) at both training and evaluation.
Point clouds with higher sampling rates are more detailed and fine-grained structures become visible at the cost of higher computational complexity.
We show in Fig. \ref{fig:pointsampling} (left) some visualizations of the influence of the sampling on the visibility of local details, which are important for both known vs unknown discrimination and closed set classification. 
In the right part Fig. \ref{fig:pointsampling}, we report the results of closed and open set performance for MSP and \eucli. 
While the closed set accuracy grows as the number of sampled points increases, there is no corresponding increase for the open set performance of \eucli.

\subsection{Synthetic to Real Benchmark}
\input{tables/tables_synth2real}
\input{tables/table_real2real}

Training on synthetic data is fundamental, especially when only a few \rw samples are available for a given task. This is often the case for 3D point cloud learning, for which ScanObjectNN \cite{scanobjectnn_ICCV19} is one of the largest publicly available \rw object datasets despite counting less than 3k samples. 
We thus analyze how models trained on synthetic data perform when tested on \rw data. 

\textbf{How does the OOD detection performance trend changes when testing on Real-World data?} 
Table \ref{tab:synth2real} provides an overview of the results on the Synthetic to Real benchmark.
W.r.t. Table \ref{tab:synth2synth} we notice a general degradation in performance due to the domain shift between train (synthetic) and test (\rw). Interestingly, the PointNet++ backbone seems to more robust to the domain shift than DGCNN. For example, the MSP baseline with PointNet++ outperforms the DGCNN counterpart by 8.9 pp in terms of AUROC and 7.4 in terms of FPR95.
We include a more comprehensive analysis of the impact of the backbone used in the Synthetic to Real benchmark in Fig. \ref{fig:scatter_backbones} (middle). 
Most of the methods that performed well in the Synthetic benchmark turn out to be less robust than the simple MSP baseline, and thus can no longer outperform it.
This is true for the vast majority of discriminative methods. For VAE it holds a similar discussion to what was done for the synthetic counterpart. NF performs consistently on both backbones with a clear AUROC improvement of 4.6 pp over MSP when applied on top of DGCNN and only a slight improvement for PointNet++. In the case of OE+mixup, the generated outliers used to finetune the classifier model are not representative of the \rw test domain and thus do not allow for improvement over the MSP baseline.

The results of distance based methods are highly dependent on the specific backbone chosen. Cosine proto performs particularly well on PN2, but fails miserably on DGCNN, most likely because DGCNN prototypes trained on synthetic are not representative of the \rw test distribution. A similar consideration can be done for \eucli, confirming the robustness of PointNet++ to the domain shift. Finally, SubArcFace demonstrates its reliability, as it achieves good results on both backbones and the best overall results on average. 

For both MSP and SubArcFace we studied the impact of several backbones also considering the closed set performance as shown in the middle part of Fig. \ref{fig:scatter_backbones}. The plot shows a linearly growing trend for both methods and the results also confirm the advantage of PointNet++ over more complex networks.

\begin{figure*}[t!]
    \centering
    \includegraphics[width=1\textwidth]{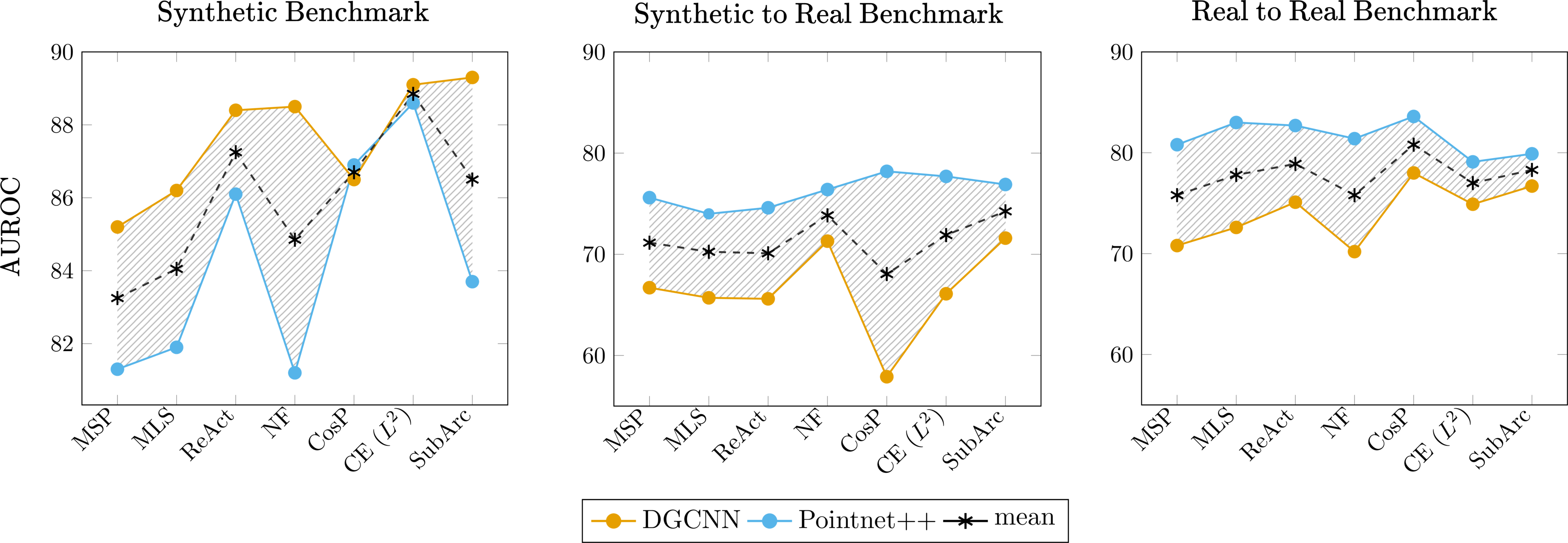}
    \caption{AUROC scores across methods and backbones for the 3DOS benchmark tracks indicated by the respective titles.\label{fig:table_plots}}
    \vspace{-4mm}
\end{figure*}

\subsection{Real to Real Benchmark}
To complete our analysis we ran the most relevant methods on the Real to Real Benchmark and present the results in Table \ref{tab:real2real}.
Overall, the trend for all approaches is consistent with what was observed in the Synthetic to Real case.
The Cosine proto approach, which already demonstrated effectiveness with PointNet++ in the Synthetic to Real benchmark, now ranks first for both DGCNN and PointNet++. 
We also highlight that PointNet++ maintains better performance than DGCNN confirming its robustness when dealing with noisy and corrupted \rw data.

For both MSP and Cosine proto we studied the impact of several backbones also considering the closed set performance as shown in the right part of Fig. \ref{fig:scatter_backbones}.

%% file: tables/tables_synth.tex
\begin{table}[t]
\caption{Results on the Synthetic Benchmark track. Each column title indicates the chosen known class set, the other two sets serve as unknown.\label{tab:synth2synth}}
\resizebox{1\textwidth}{!}{
\begin{tabular}{l@{~}c@{}cc@{}cc@{}ca@{}a}
\hline
\multicolumn{9}{c|}{Synthetic Benchmark - DGCNN~\cite{dgcnn}} \\ \hline
  &
  \multicolumn{2}{c|}{SN1 (hard)} &
  \multicolumn{2}{c|}{SN2 (med)} &
  \multicolumn{2}{c|}{SN3 (easy)} &
  \multicolumn{2}{a|}{Avg} \\

  \textit{Method} &
  {\scriptsize{AUROC}$\uparrow$ } &
  \multicolumn{1}{@{}c|}{\scriptsize{FPR95}$\downarrow$} &
  {\scriptsize{AUROC}$\uparrow$ } &
  \multicolumn{1}{@{}c|}{\scriptsize{FPR95}$\downarrow$} &
  {\scriptsize{AUROC}$\uparrow$ } &
  \multicolumn{1}{@{}c|}{\scriptsize{FPR95}$\downarrow$} &
  \multicolumn{1}{a}{\scriptsize{AUROC}$\uparrow$} &
  \multicolumn{1}{@{}a|}{\scriptsize{FPR95}$\downarrow$} \\ \hline
  
  {MSP~\cite{MSP}} &
  74.0 &
  \multicolumn{1}{@{~}c|}{83.9} &
  88.6 &
  \multicolumn{1}{@{~}c|}{62.4} &
  92.9 &
  \multicolumn{1}{@{~}c|}{43.2} &
  85.2 &
  \multicolumn{1}{@{~}a|}{63.2} \\
 
  {MLS~\cite{vaze2022openset}} &
  75.1 &
  \multicolumn{1}{@{~}c|}{77.7} &
  91.1 &
  \multicolumn{1}{@{~}c|}{42.6} &
  92.4 &
  \multicolumn{1}{@{~}c|}{35.2} &
  86.2 &
  \multicolumn{1}{@{~}a|}{51.8} \\
 
  {ODIN~\cite{ODIN}} &
  75.4 &
  \multicolumn{1}{@{~}c|}{76.5} &
  91.1 &
  \multicolumn{1}{@{~}c|}{42.9} &
  92.5 &
  \multicolumn{1}{@{~}c|}{34.4} &
  86.3 &
  \multicolumn{1}{@{~}a|}{51.3} \\
 
  {Energy~\cite{energy}} &
  75.2 &
  \multicolumn{1}{@{~}c|}{77.0} &
  91.2 &
  \multicolumn{1}{@{~}c|}{41.6} &
  92.3 &
  \multicolumn{1}{@{~}c|}{36.4} &
  86.2 &
  \multicolumn{1}{@{~}a|}{51.7} \\

  {GradNorm~\cite{gradnorm}} &
  66.2 &
  \multicolumn{1}{@{~}c|}{88.1} &
  80.9 &
  \multicolumn{1}{@{~}c|}{64.0} &
  71.6 &
  \multicolumn{1}{@{~}c|}{77.7} &
  72.9 &
  \multicolumn{1}{@{~}a|}{76.6} \\

  {ReAct~\cite{react}} &
  76.4 &
  \multicolumn{1}{@{~}c|}{74.6} &
  92.5 &
  \multicolumn{1}{@{~}c|}{37.9} &
  96.4 &
  \multicolumn{1}{@{~}c|}{19.3} &
  88.4 &
  \multicolumn{1}{@{~}a|}{{43.9}} \\ \hline

  {\vae{VAE~\cite{towards_japan}}} &
  \vae{67.2} &
  \multicolumn{1}{@{~}c|}{\vae{76.9}} &
  \vae{69.5} &
  \multicolumn{1}{@{~}c|}{\vae{83.4}} &
  \vae{94.3} &
  \multicolumn{1}{@{~}c|}{\vae{32.4}} &
  \vae{77.0} &
  \multicolumn{1}{@{~}a|}{\vae{64.2}} \\

  {NF} &
  82.0 &
  \multicolumn{1}{@{~}c|}{74.8} &
  86.1 &
  \multicolumn{1}{@{~}c|}{53.8} &
  \textbf{97.4} &
  \multicolumn{1}{@{~}c|}{\textbf{11.5}} &
  88.5 &
  \multicolumn{1}{@{~}a|}{{46.7}} \\ \hline
  
  {OE+mixup~\cite{outlier_exposure}} &
  73.7 &
  \multicolumn{1}{@{~}c|}{78.9} &
  90.4 &
  \multicolumn{1}{@{~}c|}{44.7} &
  91.4 &
  \multicolumn{1}{@{~}c|}{46.0} &
  85.2 &
  \multicolumn{1}{@{~}a|}{56.5} \\ \hline
  
  {ARPL+CS~\cite{ARPL_TPAMI}} &
  72.9 &
  \multicolumn{1}{@{~}c|}{84.2} &
  90.7 &
  \multicolumn{1}{@{~}c|}{47.1} &
  89.5 &
  \multicolumn{1}{@{~}c|}{89.5} &
  84.4 &
  \multicolumn{1}{@{~}a|}{73.6} \\
 
  {Cosine proto} &
  \textbf{84.3} &
  \multicolumn{1}{@{~}c|}{\textbf{59.1}} &
  88.8 &
  \multicolumn{1}{@{~}c|}{39.7} &
  86.4 &
  \multicolumn{1}{@{~}c|}{48.0} &
  86.5 &
  \multicolumn{1}{@{~}a|}{48.9} \\

 {\eucli} &
  80.4 &
  \multicolumn{1}{@{~}c|}{75.5} &
  90.1 &
  \multicolumn{1}{@{~}c|}{\textbf{40.9}} &
  96.7 &
  \multicolumn{1}{@{~}c|}{14.4} &
  89.1 &
  \multicolumn{1}{@{~}a|}{\textbf{43.6}} \\
 
  {SupCon~\cite{khosla2020supervised}}  &
  80.3 &
  \multicolumn{1}{@{~}c|}{75.7} &
  84.6 &
  \multicolumn{1}{@{~}c|}{73.6} &
  87.9 &
  \multicolumn{1}{@{~}c|}{44.3} &
  84.3 &
  \multicolumn{1}{@{~}a|}{64.5} \\
 
  {SubArcface~\cite{subarcface}}  &
  81.2 &
  \multicolumn{1}{@{~}c|}{73.4} &
  \textbf{91.9} &
  \multicolumn{1}{@{~}c|}{{44.0}} &
  94.9 &
  \multicolumn{1}{@{~}c|}{26.5} &
  \textbf{89.3} &
  \multicolumn{1}{@{~}a|}{48.0} \\ 
  \hline
\end{tabular}
\begin{tabular}{c@{}cc@{}cc@{}ca@{}a}
\hline
\multicolumn{8}{|c}{Synthetic Benchmark - PointNet++~\cite{pointnet++}} \\ \hline
  \multicolumn{2}{|c|}{SN1 (hard)} &
  \multicolumn{2}{c|}{SN2 (med)} &
  \multicolumn{2}{c|}{SN3 (easy)} &
  \multicolumn{2}{a}{Avg} \\

  \multicolumn{1}{|@{~}c}{\scriptsize{AUROC}$\uparrow$ } &
  \multicolumn{1}{@{}c|}{\scriptsize{FPR95}$\downarrow$} &
  {\scriptsize{AUROC}$\uparrow$ } &
  \multicolumn{1}{@{}c|}{\scriptsize{FPR95}$\downarrow$} &
  {\scriptsize{AUROC}$\uparrow$ } &
  \multicolumn{1}{@{}c|}{\scriptsize{FPR95}$\downarrow$} &
  \multicolumn{1}{a}{\scriptsize{AUROC}$\uparrow$ } &
  \multicolumn{1}{@{}a}{\scriptsize{FPR95}$\downarrow$} \\ \hline
  
  \multicolumn{1}{|@{~}c}{74.3} &
  \multicolumn{1}{@{~}c|}{82.8} &
  80.0 &
  \multicolumn{1}{@{~}c|}{78.1} &
  89.7 &
  \multicolumn{1}{@{~}c|}{52.2} &
  81.3 &
  \multicolumn{1}{@{~}a}{71.0} \\
  
  \multicolumn{1}{|@{~}c}{72.0} &
  \multicolumn{1}{@{~}c|}{80.8} &
  83.9 &
  \multicolumn{1}{@{~}c|}{64.1} &
  89.8 &
  \multicolumn{1}{@{~}c|}{40.5} &
  81.9 &
  \multicolumn{1}{@{~}a}{61.8} \\
  
  \multicolumn{1}{|@{~}c}{74.2} &
  \multicolumn{1}{@{~}c|}{79.4} &
  79.4 &
  \multicolumn{1}{@{~}c|}{71.7} &
  87.8 &
  \multicolumn{1}{@{~}c|}{41.8} &
  80.5 &
  \multicolumn{1}{@{~}a}{64.3} \\
  
  \multicolumn{1}{|@{~}c}{72.1} &
  \multicolumn{1}{@{~}c|}{81.2} &
  84.0 &
  \multicolumn{1}{@{~}c|}{64.7} &
  89.8 &
  \multicolumn{1}{@{~}c|}{39.4} &
  82.0 &
  \multicolumn{1}{@{~}a}{61.8} \\

  \multicolumn{1}{|@{~}c}{72.1} &
  \multicolumn{1}{@{~}c|}{81.8} &
  57.7 &
  \multicolumn{1}{@{~}c|}{88.9} &
  57.8 &
  \multicolumn{1}{@{~}c|}{79.0} &
  62.6 &
  \multicolumn{1}{@{~}a}{83.3} \\

  \multicolumn{1}{|@{~}c}{73.7} &
  \multicolumn{1}{@{~}c|}{79.4} &
  \textbf{89.6} &
  \multicolumn{1}{@{~}c|}{52.1} &
  \textbf{95.0} &
  \multicolumn{1}{@{~}c|}{\textbf{27.2}} &
  {86.1} &
  \multicolumn{1}{@{~}a}{{52.9}} \\ \hline
  
  \multicolumn{1}{|@{~}c}{\vae{-}} &
  \multicolumn{1}{@{~}c|}{\vae{-}} &
  \vae{-} &
  \multicolumn{1}{@{~}c|}{\vae{-}} &
  \vae{-} &
  \multicolumn{1}{@{~}c|}{\vae{-}} &
  \vae{-} &
 \multicolumn{1}{@{~}a}{ \vae{-} }\\

  \multicolumn{1}{|@{~}c}{81.5} &
  \multicolumn{1}{@{~}c|}{72.5} &
  71.1 &
  \multicolumn{1}{@{~}c|}{78.0} &
  91.0 &
  \multicolumn{1}{@{~}c|}{49.6} &
  {81.2} &
  \multicolumn{1}{@{~}a}{{66.7}} \\ \hline

  \multicolumn{1}{|@{~}c}{72.7} &
  \multicolumn{1}{@{~}c|}{78.9} &
  80.3 &
  \multicolumn{1}{@{~}c|}{68.8} &
  87.3 &
  \multicolumn{1}{@{~}c|}{62.2} &
  80.1 &
  \multicolumn{1}{@{~}a}{69.9} \\ \hline
  
  \multicolumn{1}{|@{~}c}{74.8} &
  \multicolumn{1}{@{~}c|}{80.3} &
  80.7 &
  \multicolumn{1}{@{~}c|}{72.4} &
  85.4 &
  \multicolumn{1}{@{~}c|}{50.8} &
  80.3 &
  \multicolumn{1}{@{~}a}{67.8} \\

  \multicolumn{1}{|@{~}c}{80.3} &
  \multicolumn{1}{@{~}c|}{68.3} &
  88.7 &
  \multicolumn{1}{@{~}c|}{60.8} &
  91.9 &
  \multicolumn{1}{@{~}c|}{38.0} &
  86.9 &
  \multicolumn{1}{@{~}a}{55.7} \\

  \multicolumn{1}{|@{~}c}{\textbf{83.4}} &
  \multicolumn{1}{@{~}c|}{\textbf{66.8}} &
  {89.5} &
  \multicolumn{1}{@{~}c|}{\textbf{37.7}} &
  92.9 &
  \multicolumn{1}{@{~}c|}{28.1} &
  \textbf{88.6} &
  \multicolumn{1}{@{~}a}{\textbf{44.2}} \\

  \multicolumn{1}{|@{~}c}{80.9} &
  \multicolumn{1}{@{~}c|}{75.5} &
  83.5 &
  \multicolumn{1}{@{~}c|}{68.2} &
  85.1 &
  \multicolumn{1}{@{~}c|}{45.1} &
  83.2 &
  \multicolumn{1}{@{~}a}{62.9} \\

  \multicolumn{1}{|@{~}c}{79.0} &
  \multicolumn{1}{@{~}c|}{81.2} &
  82.9 &
  \multicolumn{1}{@{~}c|}{60.3} &
  89.1 &
  \multicolumn{1}{@{~}c|}{32.8} &
  {83.7} &
  \multicolumn{1}{@{~}a}{58.1} \\ 
  \hline
\end{tabular}
}\vspace{-3mm}
\end{table}

\begin{table}[t!]
\caption{Relationship between closed and open set performance when training a discriminative model via the addition of Label Smoothing (LS). We show the results on the hard SN1 set.\label{tab:LS}}
\resizebox{1\textwidth}{!}{
\begin{tabular}{cc@{~~}c|c@{~~}c|c@{~~}c||c@{~~}c|}
\hline
\multicolumn{9}{c|}{SN1 (hard) -  Synthetic Benchmark - DGCNN~\cite{dgcnn}}\\
\hline
\multirow{2}{*}{} & \multirow{2}{*}{\small{MSP}} & \multirow{2}{*}{\small{+LS}} &  \multirow{2}{*}{\small{MLS}} & \multirow{2}{*}{\small{+LS}} &  \multirow{2}{*}{\small{\eucli}} & \multirow{2}{*}{\small{+LS}} & \multicolumn{2}{c|}{\small{Closed set}} \\
 &  &  &   &  &   &  &  {\small{Acc}} & {\small{+LS}}\\
\hline
\small{{AUROC}$\uparrow$} & 74.0 & 77.4 & 75.1 & 77.5 & 80.4 & 80.2 &  \multirow{2}{*}{85.5} &\multirow{2}{*}{86.0}\\
\small{{FPR95}$\downarrow$} & 83.9 & 73.7 & 77.7 & 71.8 & 75.5 & 66.7 &  &\\
\hline
\end{tabular}
\begin{tabular}{|c@{~~}c|c@{~~}c|c@{~~}c||c@{~~}c}
\hline
\multicolumn{8}{|c}{SN1 (hard) -  Synthetic Benchmark - PointNet++~\cite{pointnet++}}\\
\hline
\multirow{2}{*}{\small{MSP}} & \multirow{2}{*}{\small{MSP+LS}} &  \multirow{2}{*}{\small{MLS}} & \multirow{2}{*}{\small{MLS+LS}} &  \multirow{2}{*}{\small{\eucli}} & \multirow{2}{*}{\small{+LS}} &  \multicolumn{2}{c}{\small{Closed set}} \\
&  &   &  &   &  &  {\small{Acc}} & {\small{+LS}}\\
 \hline
 74.3 & 72.7 & 72.0 & 69.6 & 83.4 & 79.1 &  \multirow{2}{*}{85.9} &\multirow{2}{*}{86.3} \\
 82.8 & 78.6 & 80.8 & 77.6 & 66.8 & 77.4 &  &\\
 \hline
\end{tabular}
}
\vspace{-4mm}
\end{table}

\begin{figure}
    \centering
    \includegraphics[width=1\textwidth]{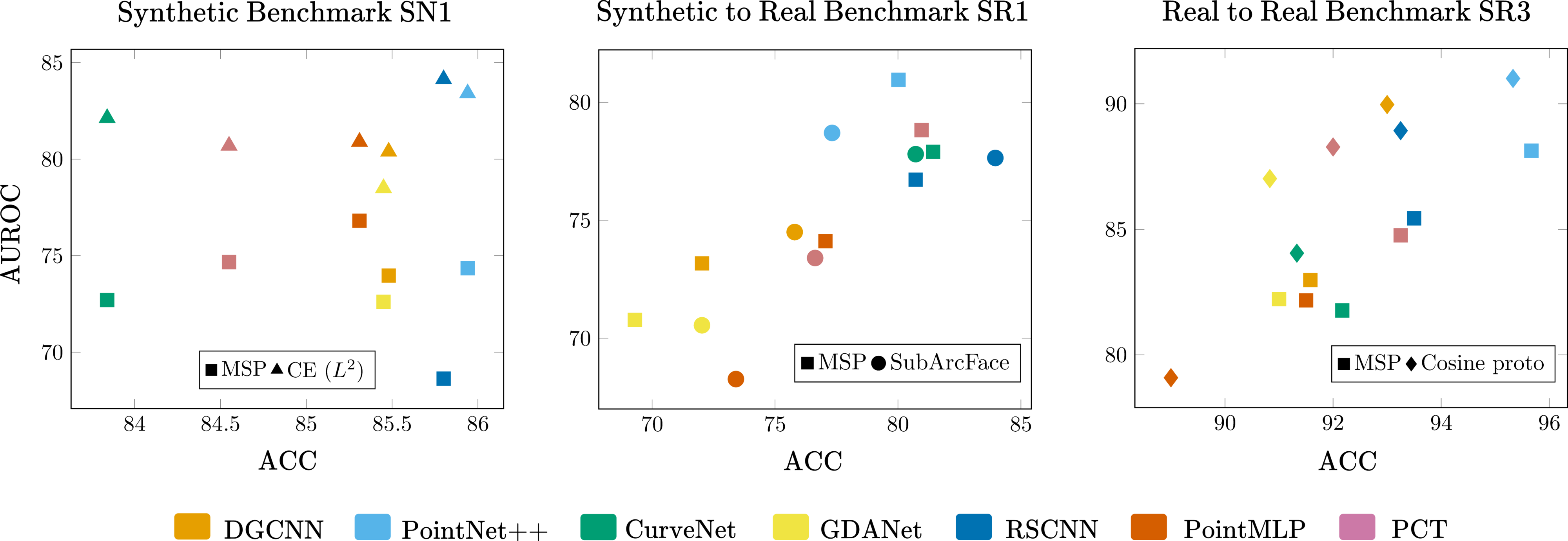}
    \caption{Correlation between AUROC and ACC performance when changing the backbone on the Synthetic Benchmark SN1 case (left), the Synthetic to Real Benchmark SR1 case (middle) and the Real to Real Benchmark SR3 case (right).\label{fig:scatter_backbones}}
    \vspace{-4mm}
\end{figure}

%% file: tables/tables_synth2real.tex
\begin{table}[t]
\caption{Results on the Synthetic to Real Benchmark track. Each column title indicates the chosen known class set, the other two sets serve as unknown.\label{tab:synth2real}}
\resizebox{1\textwidth}{!}{
    \begin{tabular}{l@{~}c@{~~}c|c@{~~}c|a@{~~}a|}
    \hline
    \multicolumn{7}{c|}{Synth to Real Benchmark - DGCNN~\cite{dgcnn}}\\ \hline
     & \multicolumn{2}{c|}{SR 1 (easy)}
    & \multicolumn{2}{c|}{SR 2 (hard)} &
    \multicolumn{2}{a|}{Avg}\\
    \emph{Method} &
      {\scriptsize{AUROC}$\uparrow$} &
      {\scriptsize{FPR95}$\downarrow$} &
      {\scriptsize{AUROC}$\uparrow$} &
      {\scriptsize{FPR95}$\downarrow$} &
      {\scriptsize{AUROC}$\uparrow$} &
      {\scriptsize{FPR95}$\downarrow$} \\ \hline
    MSP~\cite{MSP} & 72.2 & {91.0} & 61.2 & {90.3} & {66.7} & 90.6 \\
    MLS & 69.0 & {92.2} & 62.4 & {88.9} & 65.7 & 90.5 \\
    ODIN~\cite{ODIN} & 69.0 & {92.2} & 62.4 & {89.0} & 65.7 & 90.6 \\
    Energy~\cite{energy} & 68.8 & {92.7} & 62.4 & {88.9} & 65.6 & 90.8 \\
    GradNorm~\cite{gradnorm} & 67.0 & {93.5} & 59.8 & {89.4} & 63.4 & 91.5 \\
    ReAct~\cite{react} & 68.4 & {92.1} & 62.8 & {88.8} & 65.6 & {90.5} \\
    \hline
    \vae{VAE}~\cite{towards_japan} & \vae{68.6} & {\vae{77.0}} & \vae{57.9} & {\vae{92.3}} & \vae{63.3} & \vae{84.6} \\
    NF  & 72.5 & {\textbf{81.6}} & \textbf{70.2} & {\textbf{83.0}} & {71.3} & \textbf{82.3} \\
    \hline
    OE+mixup~\cite{outlier_exposure} & 71.1 & {89.6} & 59.5 & {92.0} & 65.3 & 90.8 \\
    \hline
    ARPL+CS~\cite{ARPL_TPAMI} & 71.5 & {90.2} & 62.8 & {89.5} & 67.1 & 89.8 \\
    Cosine proto & 58.6 & {90.6} & 57.3 & {91.3} & 57.9 & 91.0 \\
    \eucli & 67.5 & {87.4} & 64.6 & {91.0} & 66.1 & 89.2 \\
    SubArcFace~\cite{subarcface} & \textbf{74.5} & {86.7} & 68.7 & {86.6} & \textbf{71.6} & {86.7} \\
    \hline
    
    \end{tabular}
    \begin{tabular}{|c@{~~}c|c@{~~}c|a@{~~}a}
    \hline
    \multicolumn{6}{|c}{Synth to Real Benchmark - PointNet++~\cite{pointnet++}}                          \\ \hline
    \multicolumn{2}{|c|}{SR 1 (easy)}  &
    \multicolumn{2}{c|}{SR 2 (hard)}  & \multicolumn{2}{a}{Avg}       \\
    {
    \scriptsize{AUROC} $\uparrow$} &
    {\scriptsize{FPR95} $\downarrow$} &
    {\scriptsize{AUROC} $\uparrow$} &
    {\scriptsize{FPR95} $\downarrow$} &
    {\scriptsize{AUROC} $\uparrow$} &
    {\scriptsize{FPR95} $\downarrow$} \\ 
        \hline
        81.0 & {79.6} & 70.3 & {86.7} & 75.6 & 83.2 \\
        82.1 & {76.6} & 67.6 & {86.8} & 74.8 & 81.7 \\
        81.7 & {77.3} & 70.2 & {84.4} & {76.0} & {80.8} \\
        81.9 & {77.5} & 67.7 & {87.3} & 74.8 & 82.4 \\
        77.6 & {80.1} & 68.4 & {86.3} & 73.0 & 83.2 \\
        81.7 & {75.6} & 67.6 & {87.2} & 74.6 & 81.4 \\
        \hline
        -    & {-}    & -    & {-}    & -   & - \\
        78.0 & {84.4} & 74.7 & {84.2} & {76.4} & {84.3} \\
        \hline
        71.2 & {89.7} & 60.3 & {93.5} & 65.7 & 91.6 \\
        \hline
        \textbf{82.8} & {74.9} & 68.0 & {89.3} & 75.4 & 82.1 \\
        79.9 & {\textbf{74.5}} & \textbf{76.5} & {\textbf{77.8}} & \textbf{78.2} & \textbf{76.1}  \\
        79.7 & {84.5} & 75.7 & {80.2} & 77.7 & 82.3 \\
        78.7 & {84.3} & 75.1 & {83.4} & 76.9 & 83.8 \\
        \hline
    \end{tabular}
}
\end{table}

%% file: tables/table_real2real.tex
\begin{table}[t]
\caption{Results on the Real to Real Benchmark track. Each column title indicates the chosen unknown class set, the other two sets serve as known.\label{tab:real2real}}
\resizebox{1\textwidth}{!}{
\begin{tabular}{l@{~}c@{}c|c@{}c|c@{}c|a@{~~}a|}
\hline
\multicolumn{9}{c|}{Real to Real Benchmark - DGCNN~\cite{dgcnn}} \\ \hline
  &
  \multicolumn{2}{c|}{SR3 (easy)} &
  \multicolumn{2}{c|}{SR2 (med)} &
  \multicolumn{2}{c|}{SR1 (hard)} &
  \multicolumn{2}{a|}{Avg} \\
  \textit{Method} &
  {\scriptsize{AUROC}$\uparrow$ } &
  {\scriptsize{FPR95}$\downarrow$} &
  {\scriptsize{AUROC}$\uparrow$ } &
  {\scriptsize{FPR95}$\downarrow$} &
  {\scriptsize{AUROC}$\uparrow$ } &
  {\scriptsize{FPR95}$\downarrow$} &
  {\scriptsize{AUROC}$\uparrow$} &
  {\scriptsize{FPR95}$\downarrow$} \\ \hline
    {MSP~\cite{MSP}} &
    83.0 & 69.4 & 72.0 &	88.7 & 57.5	& 90.3 & 70.8 & 82.8 \\
    {MLS~\cite{vaze2022openset}} &
    84.9 & 58.2 & 79.0 & 81.0 & 54.0 & 92.8 & 72.6 & 77.3 \\
    {ODIN~\cite{ODIN}} &
    84.9	& 58.2 & 79.0 & 80.9 & 54.0 & 92.8 & 72.6 & 77.3 \\
    {Energy~\cite{energy}} &
    84.8 & 59.7 &	79.1 &	81.4  &	53.8  &	93.2  &	72.6  &	78.1 \\
    {GradNorm~\cite{gradnorm}} &
    77.5	& 73.3 & 73.3 & 87.4 & 51.0 & 92.9 & 67.2 & 84.5 \\
    {ReAct~\cite{react}} &
    87.6	& 54.0 & 79.0 & 78.6 & 58.9 & 93.1 & 75.1 & 75.3 \\
    \hline
    {NF} &
    76.9 & 77.3 & 71.7 & 82.7 & 61.8 & 86.2 & 70.2 & 82.1 \\
    \hline
    {OE+mixup~\cite{outlier_exposure}} &
    76.8 & 77.8 & 74.9 & 87.2 & 57.6 & 89.9 & 69.8 & 85.0 \\
    \hline
    {Cosine proto} &
    \textbf{90.0} & \textbf{43.7} & \textbf{78.5} & \textbf{75.3} & 65.5 & 85.7 & \textbf{78.0} & \textbf{68.2} \\
    {\eucli} &
    83.1 & 59.3 & 74.5 & 77.2 & \textbf{67.1} & 86.8 & 74.9 & 74.4  \\
    {SubArcface~\cite{subarcface}}  &
    86.7 & 58.5 & 78.4 & 76.1 & 65.0 & \textbf{84.0} & 76.7 & 72.9 \\ 
    \hline
\end{tabular}
%
\begin{tabular}{|c@{}c|c@{}c|c@{}c|a@{~~}a}
\hline
\multicolumn{8}{|c}{Real to Real Benchmark - PointNet++~\cite{pointnet++}} \\ \hline
  \multicolumn{2}{|c|}{SR3 (easy)} &
  \multicolumn{2}{c|}{SR2 (med)} &
  \multicolumn{2}{c|}{SR1 (hard)} &
  \multicolumn{2}{a}{Avg} \\
  {\scriptsize{AUROC}$\uparrow$ } &
  {\scriptsize{FPR95}$\downarrow$} &
  {\scriptsize{AUROC}$\uparrow$ } &
  {\scriptsize{FPR95}$\downarrow$} &
  {\scriptsize{AUROC}$\uparrow$ } &
  {\scriptsize{FPR95}$\downarrow$} &
  {\scriptsize{AUROC}$\uparrow$ } &
  {\scriptsize{FPR95}$\downarrow$} \\ \hline
    88.1 & 67.3 & 80.6 & 84.0 & 73.7 & 80.3 & 80.8 & 77.2 \\
    89.4 & 53.8 & \textbf{83.4} & 73.1 & 76.4 & \textbf{75.3} & 83.0 & 67.4 \\
    90.2 & 47.9 & 83.3	& 71.7 & 76.3 & 76.8 & 83.3	& 65.5 \\
    89.5 & 50.6 & 81.6	& 75.8 & 76.6 & 75.5 & 82.6	& 67.3 \\
    88.5 & 50.7 & 77.4	& 75.3 & 75.2 & 76. 8 & 80.4 & 67.6 \\
    90.3 & 48.9 & 82.4	& 75.8 & 75.4 & 77.6 & 82.7 & 67.4 \\ \hline
    88.0 & 47.7 & 80.6	& \textbf{68.2} & 75.6 & 81.4 & 81.4 & 65.8 \\ \hline
    72.6 & 83.5 & 72.0 & 88.5 & 62.5 & 87.8 & 69.0 & 86.6 \\ \hline
    \textbf{91.0} & \textbf{41.0} & 82.1 & 78.2 & \textbf{77.6} & 75.6 & \textbf{83.6} & \textbf{64.9} \\
    85.1 & 64.4 & 78.9 & 83.9	& 73.2 & 79.1 & 79.1 & 75.8 \\
    87.1 & 61.3 & 78.9 & 76.9	& 73.7 & 81.4 & 79.9 & 73.2 \\ 
  \hline 
\end{tabular}
}
\end{table}

%% file: sections/conclusion.tex
We presented 3DOS, the first benchmark for 3D Open Set learning that takes into account several settings and three scenarios with different types of distributional shifts. 
Our analysis reveals that cutting-edge 2D Open Set methods do not easily transfer their state-of-the-art performance to 3D data, with simple representation learning approaches such as \eucli,  SubArcFace and Cosine proto often outperforming them. 
Furthermore, the performances of the 3D Open Set methods depend on the chosen backbone: PointNet++ has proven to be extremely robust in processing \rw data, even across domains, outperforming more recent and complex networks.
The point density may be an issue for baseline approaches but has a minimal impact on distance-based strategies as \eucli.
Finally, Open Set on 3D data becomes extremely difficult when dealing with the combination of semantic and domain shift. 

Figure \ref{fig:table_plots} depicts a summary overview of the three studied scenarios indicating how the Synthetic to Real is the most challenging case, followed by the Real to Real and finally by the Synthetic Benchmark. This confirms that the domain shift between synthetic and real data adds extra challenges over the semantic shift. 
Moreover, it is interesting to notice that the improvement provided by the best Open Set methods over the MLS/MSP baselines is quite visible in the Synthetic Benchmark (DGCNN SubArcFace > MLS, +3.1 AUROC), but is reduced in the Synthetic to Real (PointNet++ Cosine Proto > MSP, + 2.6 AUROC) and Real to Real cases (PointNet++ Cosine Proto > MLP, + 0.6 AUROC), which clearly asks for new approaches and reveals room for improvements.

We hope that this benchmark will serve as a solid foundation for future research in this area, pushing for the development of Open Set methods tailored for 3D data and able to exploit their peculiarity.
\paragraph{Acknowledgements} We acknowledge the CINECA award under the ISCRA initiative, for the availability of high performance computing resources and support. We also acknowledge the support of the European H2020 Elise project (\url{www.elise-ai.eu}).

%% file: sections_suppl/supp-details.tex
\section{Baselines details, implementation and reproducibility}

We publicly release our code and data at \url{https://github.com/antoalli/3D_OS}. The repository also contains instructions on how to replicate all the experiments.

In the main paper we include a high-level description of all methods and the most relevant implementation details. Here we extend the description, discussing implementation choices and hyperparameters for each.

\textbf{Discriminative Models} 

All the approaches in this group (MSP, MLS, ODIN, Energy, GradNorm, ReAct) share exactly the same basic cross-entropy classifier trained on known data. 
Section 4.1 of the main paper already specified the cardinality (number of points) of the point clouds in train and test, as well as the number of epochs, initial learning rate and learning rate scheduling policy.

\emph{MSP \& MLS} only differ for the way in which the logits of the classifier on each test sample are used to compute the normality score. MLS directly employs the maximum logit, while for MSP the logits go through a softmax function before selecting the maximum of the obtained class probabilities as the score.

\emph{ODIN} internally exploits input perturbation and temperature scaling since both have an effect on the distribution of the softmax scores, better separating data from known and unknown classes. 
The most important hyperparameter here is the temperature value which we set to $T=1000$ following the original paper's instructions~\cite{ODIN}. The input perturbation magnitude $\varepsilon$ should be optimized through a validation set of OOD samples, still keeping its value very small to avoid detrimental effects. Considering that we do not have access to OOD data at training time we preferred to stay on the safe side, setting $\varepsilon=0$ in all of our experiments, effectively disabling the input perturbation. 

\emph{Energy}
The energy-based normality score is computed by postprocessing the network output logits and it's hyperparameter free. 

\emph{GradNorm}
In order to compute a normality score, the KL divergence between the network output and a uniform distribution is backpropagated to obtain network gradients and then extract their norm. 
As suggested in the original paper~\cite{gradnorm}, we exploit the norm of the gradients of the last layer only. No further hyperparameters are involved in this process.

\emph{ReAct} rectifies the test-time activations of the network trained for classification and then can exploit any normality score computation strategy.
By following~\cite{react} we use ReAct in combination with Energy normality score. The only hyperparameter involved is the rectification threshold value.
We chose it by exploiting the known class validation samples to preserve 90\% of ID activations.
For the Synthetic Benchmark the known classes validation samples come from the original ShapeNetCore~\cite{shapenet_dataset} validation split. 
For the Synthetic to Real Benchmark, we adopt ModelNet40~\cite{mn40_dataset} known classes test set as validation, we underline that these samples are not involved in the testing phase since both Closed Set accuracy and Open Set performance are computed on ScanObjectNN~\cite{scanobjectnn_ICCV19}.

\textbf{Density and Reconstruction Based Models} 

\emph{VAE} For our experiments we use the original code publicly released by the authors\footnote{\url{https://github.com/llien30/point_cloud_anomaly_detection}} \cite{towards_japan}, as well as their same choice on point cloud cardinality (2048 points) and  hyperparameters.
The encoder is composed of graph-convolutional layers: it takes as input a point cloud and outputs two 512-dimensional vectors representing the mean and variance. 
The decoder is a FoldingNet: it takes in input a sampled vector z from the encoded mean and variance and outputs an intermediate and a final point cloud reconstruction with respectively 1024 and 2048 points.
The normality score is computed as the Chamfer Distance between the original test sample and its final reconstruction. 

\emph{NF} 
For this method we got inspired by~\cite{OpenHybrid}. The overall architecture consists of three modules: a feature encoder, a classification head, and a Normalizing Flow (NF)  head.
The feature encoder and classification head work together to optimize a standard cross-entropy loss.
The NF head works independently on top of the feature encoder representation and it is composed of eight Real-NVP~\cite{realnvp} coupling blocks which are trained to maximize the log-likelihood of the observed training features.
At inference time we use the test sample log-likelihood as a normality score. For training NF we use the Adam optimizer with a learning rate of 0.0002 and weight decay set to 0.00001.

\textbf{Outlier Exposure with OOD Generated Data}
The outlier exposure strategy described in~\cite{outlier_exposure} consists in training a Discriminative Model on ID training data (known) through standard cross-entropy loss, while exploiting additional OOD training data (unknown) to improve ID-OOD separability.
Specifically, we start from the same cross-entropy classifier trained on known classes employed for the first group of strategies (i.e. Discriminative Models) and finetune it by minimizing the following loss function: $\mathcal{L}_{ft} = \mathcal{L}_{CE, known} + \lambda \mathcal{L}_{OE, unknown}$. 

The finetuning involves a continued optimization of the cross-entropy loss on known training data $\mathcal{L}_{CE, known}$ and an outlier exposure objective $\mathcal{L}_{OE, unknown}$ on unknown training data.
The goal of the outlier exposure objective is to minimize the KL divergence between the Uniform and Cross-entropy distributions for unknown samples.
The hyperparameter $\lambda$ controls the importance of the OE auxiliary objective and is set to $0.5$ according to the original paper~\cite{outlier_exposure}. The finetuning is performed for additional 100 epochs, with a learning rate reduced by a factor of 100. 

For OE finetuning, OOD training data are obtained through Rigid Subset Mix (RSMix)~\cite{Lee_2021_CVPR} of known class samples, some examples of the produced OOD data for the synthetic SN1 set are shown in Fig. \ref{fig:rsmix_examples}.

\textbf{Representation and distance-based methods} 

\emph{ARPL+CS} The training process of this method
involves learning a GAN model designed to generate confusing samples (CS), as well as optimizing the reciprocal points learning objective.
The model needs a number of hyperparameters to keep all the learning components well-balanced and we used the same values adopted by the authors~\cite{ARPL_TPAMI}.

\emph{Cosine proto} We train a simple cosine classifier by following CosFace~\cite{cosface} strategy and setting the imposed margin to 0. At inference time output logits correspond to the cosine similarities between the test sample and the class prototypes. The largest value is used as the normality score without introducing additional hyperparameters. 

\emph{\eucli} With this method we aim at studying the reliability to OOD detection of the feature representation learned by the same classifier trained for the Discriminative Models.
We use the inverse of the distance from the nearest training sample as each test sample normality score, without introducing additional hyperparameters.

\emph{SupCon} requires long training with a large batch size to reach convergence~\cite{khosla2020supervised}. With respect to the models trained for classification we double the batch size and halve the learning rate. To deal with large batches we perform distributed training on multiple GPUs and adopt Synchronized Batch Normalization. 
We also increase the number of epochs to 2000, using a linear warmup in the first 100. During deployment the normality score is the cosine similarity of each test sample to its nearest training sample.
The SupCon learning objective builds a hyperspherical feature space in which class clusters are compact and well separated. Similar results can be obtained through SubArcFace which exploits a much more easily optimized classification-like loss.
On this basis, and also considering the poor results of SupCon  on the Synthetic Benchmark, we decided to discard it in the Synthetic to Real and Real to Real Benchmarks. 

\emph{SubArcFace} learning objective seeks to maximize the cosine similarity between each training sample and one of the $K$ centers associated with its respective class, while also imposing a certain margin $m$ between different classes. 
We use $K=3$ as done by the authors in the original paper~\cite{subarcface} and set the margin to $m=0.5$, as done in ArcFace~\cite{arcface} from which this hyperparameter is inherited. The normality score is computed as done for SupCon. 

%% file: sections_suppl/supp-synth-real.tex
\section{Synthetic to Real Benchmark: Additional Analyses}
The goal of the Synthetic to Real Benchmark track is to simulate realistic deployment conditions and analyze the behaviour of Open Set methods in this context. 
Indeed, due to the high cost of 3D data acquisition and labelling, large synthetic datasets are commonly used to train deep neural network models
which are then employed in real-world applications such as autonomous driving, augmented reality or robotics.

\begin{figure*}[t!]
    \centering
    \includegraphics[width=1\textwidth]{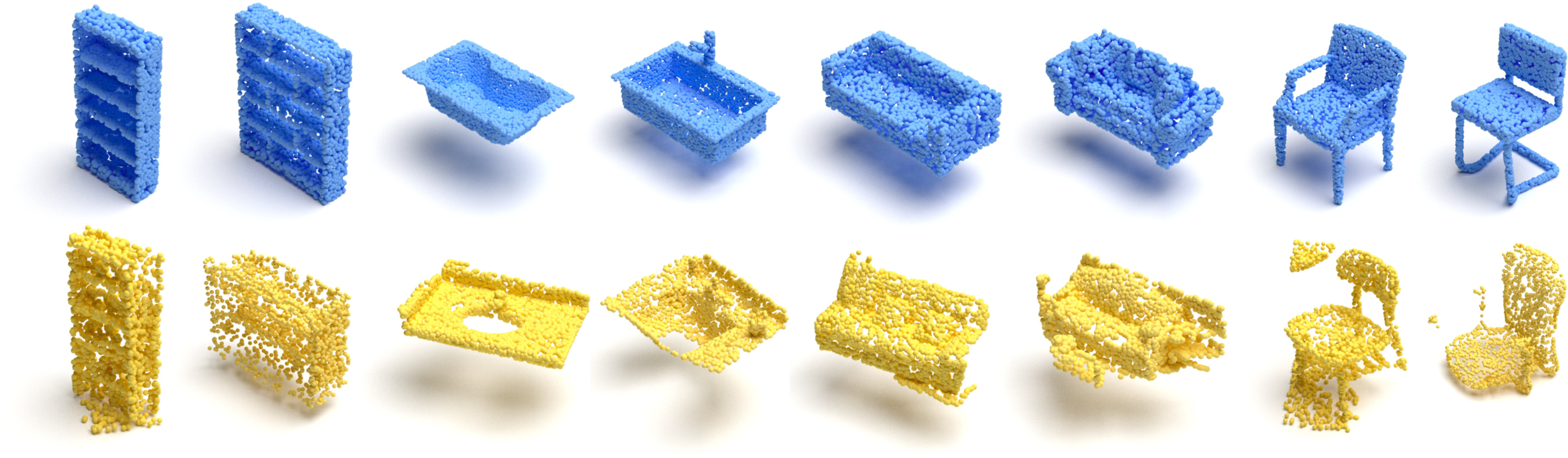}
    \caption{Qualitative visualizations: point clouds from bookshelf, sink, sofa and chair categories. Blue are synthetic point clouds from ModelNet40, yellow are \rw from ScanObjectNN
    \label{fig:synth2real_renders}}
\end{figure*}

This strategy, although effective in lowering data collection costs, inevitably causes a covariate distribution (visual domain) shift between training and test data.
As a result test samples belonging to unknown classes show both a semantic and a domain shift, while test samples belonging to known classes only show a domain shift. The necessity to distinguish between these two cases raises the difficulty of the unknown detection task.
To get an idea of the difference between the synthetic and the real domains  we render some point clouds in Figure \ref{fig:synth2real_renders}, respectively from the known classes in the train and test of our Synthetic to Real Benchmark track. 

It is evident that \rw samples (in yellow) are much noisier than synthetic ones (in blue). Moreover, \rw point clouds have background (first chair), are affected by occlusion, partiality (second chair) and interaction with other objects nearby (second sofa and first chair). 

This additional covariate shift, which is present only in the Synthetic to Real Benchmark, is what makes this track the most difficult among the ones we analyzed.

\subsection{AUPR metric}
\input{tables_suppl/table_synth_real_AUPR}

In the results reported in the main paper we exploited two of the most common OOD detection metrics to evaluate the unknown detection ability of the analyzed methods: AUROC and FPR95. Different metrics may be chosen for the same purpose, one of the most used being the Area Under the Precision Recall curve (\textbf{AUPR}). Similarly to AUROC this is a threshold-free metric: the $\textit{precision} = TP/(TP+FP)$, is plotted as a function of $\textit{recall} = TP/(TP+FN)$, for different threshold settings and then the area under the resulting curve is computed, with a high value (near to 1) indicating a good known-unknown separation and a low one (near to 0) highlighting bad performance.
Differently from the AUROC, the AUPR takes care of the possible unbalance between positive and negative classes by adjusting for the base rates.
Specifically, we computed the AUPR considering the unknown samples as positive: given its complementary polarity with AUROC we expect it to provide further information. 

We report the results of the best performing backbone on the most difficult benchmark track, i.e. we run with PointNet++~\cite{pointnet++} on the Synthetic to Real Benchmark: see Table \ref{tab:synth2real_AUPR}. 
According to AUPR, the top performing methods are the same ones that AUROC and FPR95 highlighted as best. 

\subsection{OE+mixup in the Synthetic to Real Benchmark}

Figure \ref{fig:rsmix_examples} presents point cloud instances obtained via mixup. This strategy is used to create data that can be exploited as OOD during training via outlier exposure. However, as it can be noticed, shape mixing inevitably introduces some artefacts that resemble noise, missing parts and background, typical of \rw data. We believe that in the synthetic-to-real experiments this may introduce some confusion rather than helping in separating known and unknown classes, as it pushes the model to believe that all corrupted samples belong to unknown classes. This reflects in the poor \emph{OE+mixup} results reported in Tab. \ref{tab:synth2real}. 

\subsection{Corruption-based data augmentation}

\input{tables_suppl/table_synth_real_corruptions}

The results of the Synthetic to Real benchmark highlight the impact of the domain shift on the open set performance. Indeed, synthetic point clouds exhibit a clean geometry and have no background. Differently, real-world point clouds are affected by partiality, and occlusion, cluttered with noise and background points.
A possible solution to partially bridge this domain shift consists in trying to emulate these kinds of corruptions at training time via tailored data augmentation functions. 

\begin{figure*}[t!]
    \begin{minipage}{0.48\textwidth}
        \centering
        \vspace{8mm}
        \includegraphics[width=\textwidth]{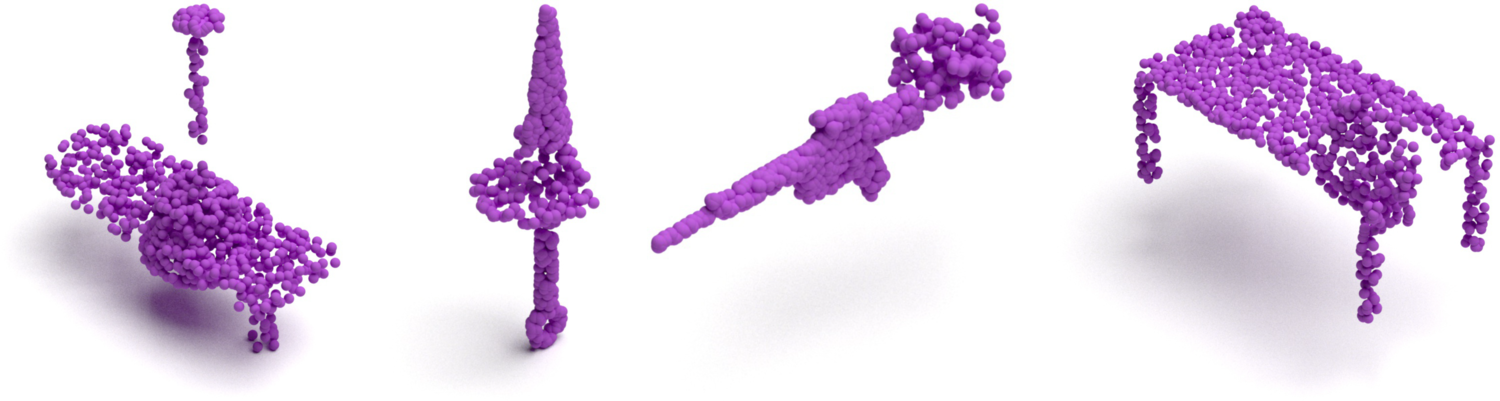}
        \caption{Examples of RSMix~\cite{Lee_2021_CVPR} between known samples of the synthetic SN1 set. We employ these mixed point clouds as training OOD data in OE experiments \label{fig:rsmix_examples}}
    \end{minipage}
    \hspace{2mm}
    \centering
        \begin{minipage}{0.48\textwidth}
        \centering
        \includegraphics[width=\textwidth]{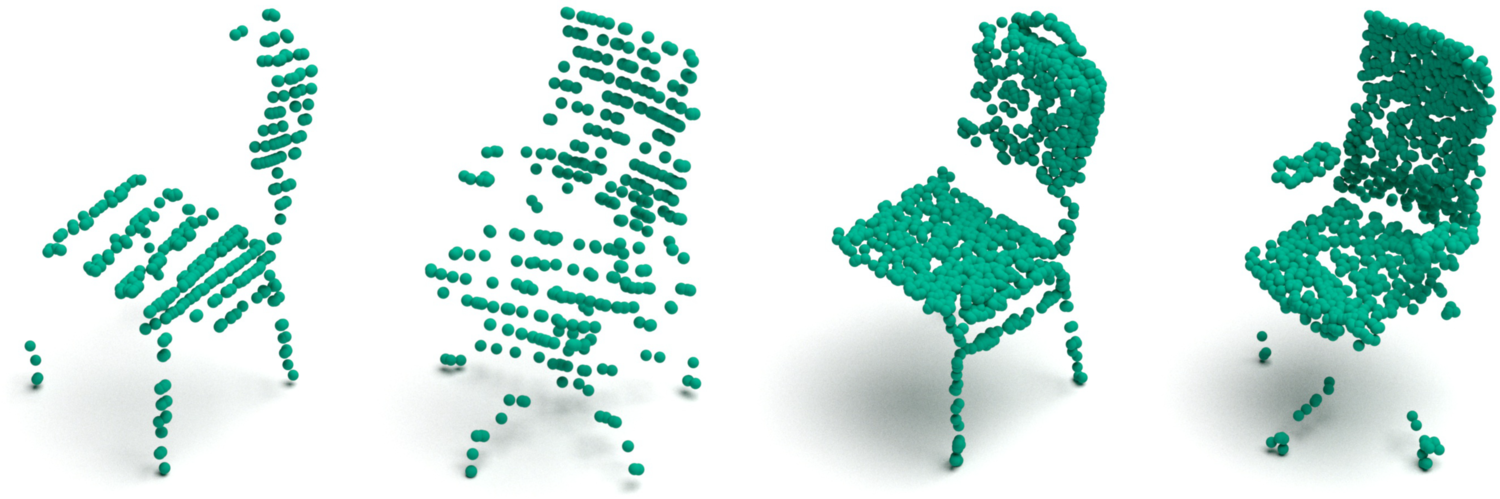}
        \caption{Examples of ModelNet point clouds augmented with LIDAR (first two) and Occlusion (last two) corruptions from \cite{corruptions} \label{fig:corruption_examples}}
    \end{minipage}
    \hfill
\end{figure*}

We experiment with this solution by adopting the \emph{Occlusion} and \emph{LIDAR} augmentations from \cite{corruptions}. Fig. \ref{fig:corruption_examples} shows some examples of the results obtained using these transformations. Comparing them with point clouds in the second row of Fig. \ref{fig:synth2real_renders} it is possible to notice some differences. 
Considering both DGCNN and PointNet++ backbones we perform experiments with such augmented training data for both the simple MSP baseline and SubArcFace method.
Results for these experiments are presented in Tab.~\ref{tab:synth2real_corr} where we refer to the models trained with the augmented training set as (+RW Augm).
DGCNN highly benefit from the \rw tailored data augmentation, obtaining an AUROC improvement of +7.2pp and +3.5pp respectively for MSP and SubArcFace methods. 
PointNet++, on the other hand, has already proven its robustness in synthetic to \rw scenario and does not benefit from the tailored augmentation schema.

\subsection{Number of points at test time}
\input{tables_suppl/table_test_time_n_points}

The visual domain shift between training and test conditions that appear in the Synthetic to Real case may include also a difference in the cardinality of points. 
In our Synthetic to Real Benchmark we followed the same procedure adopted for the Synthetic Benchmark case: we use 1024-dim synthetic points clouds during training, with points randomly sampled from the surface of the ModelNet40 meshes.
The trained model is then evaluated solely on \rw samples from the ScanObject dataset. Real-world data samples come directly in the form of 2048-dim point clouds. For each sample, out of the 2048 points, we do not know a priori which are belonging to the foreground object, background, or other interacting objects.
In order to ease results replication, we originally decided to avoid random subsampling and used at test time 2048-dim \rw points clouds in all our Synthetic to Real experiments. 
We are now interested in analyzing how the number of points used during inference influences the results and thus we test with 1024 and 512 and report results in Tab. \ref{tab:test_n_points}. 
Looking at the results we can conclude that forcing train and test data to have the same number of points (1024-1024) slightly reduces the domain shift and provides a small performance improvement with respect to the remaining cases that present an asymmetry in the point cloud cardinality (1024-2048, 1024-512).

%% file: tables_suppl/table_synth_real_AUPR.tex
\begin{table}[t]
\caption{AUROC, FPR95, AUPR results on the Synthetic to Real Benchmark with PointNet++~\cite{pointnet++} \label{tab:synth2real_AUPR}}
\resizebox{1.0\textwidth}{!}{
    \begin{tabular}{l@{~}c@{~~}c@{~~}c|c@{~~}c@{~~}c|a@{~~}a@{~~}a}
    \hline
    \multicolumn{10}{c}{Synthetic to Real Benchmark - PointNet++~ ~\cite{pointnet++}}\\
    \hline
    & \multicolumn{3}{c|}{SR 1}  &
    \multicolumn{3}{c|}{SR 2}  & \multicolumn{3}{a}{Avg} \\
    \textit{Method} & {\scriptsize{AUROC} $\uparrow$} &
      {\scriptsize{FPR95} $\downarrow$} & {\scriptsize{AUPR} $\uparrow$} &
      {\scriptsize{AUROC} $\uparrow$} &
      {\scriptsize{FPR95} $\downarrow$} & {\scriptsize{AUPR} $\uparrow$} &
      {\scriptsize{AUROC} $\uparrow$} &
      {\scriptsize{FPR95} $\downarrow$} & {\scriptsize{AUPR} $\uparrow$} \\ \hline
    MSP [18]        & 81.0 & 79.6 & 79.9 & 70.3 & 86.7 & 83.9 & 75.6 & 83.2 & 81.9 \\
    MLS [44]        & 82.1 & 76.6 & 82.0 & 67.6 & 86.8 & 83.1 & 74.8 & 81.7 & 82.5 \\
    ODIN [27]       & 81.7 & 77.3 & 81.5 & 70.2 & 84.4 & 84.4 & 76.0 & 80.8 & 82.9 \\
    Energy [28]     & 81.9 & 77.5 & 82.0 & 67.7 & 87.3 & 83.0 & 74.8 & 82.4 & 82.5 \\
    GradNorm [21]   & 77.6 & 80.1 & 78.8 & 68.4 & 86.3 & 83.5 & 73.0 & 83.2 & 81.2 \\
    ReAct [40]      & 81.7 & 75.6 & 81.9 & 67.6 & 87.2 & 83.1 & 74.6 & 81.4 & 82.5 \\ \hline
    NF              & 78.0 & 84.4 & 77.0 & 74.7 & 84.2 & 86.3 & 76.4 & 84.3 & 81.7 \\ \hline
    OE+mixup [19]   & 71.2 & 89.7 & 70.9 & 60.3 & 93.5 & 77.9 & 65.7 & 91.6 & 74.4 \\ \hline
    ARPL+CS [7]     & \textbf{82.8} & {74.9} & \textbf{82.7} & 68.0 & 89.3 & 83.4 & 75.4 & 82.1 & 83.0 \\
    Cosine proto    & 79.9 & {\textbf{74.5}} & 81.2 & \textbf{76.5} & {\textbf{77.8}} & \textbf{88.1} & \textbf{78.2} & \textbf{76.1} & \textbf{84.7} \\
    \eucli          & 79.7 & 84.5 & 78.4 & 75.7 & 80.2 & 87.3 & 77.7 & 82.3 & 82.9 \\
    SubArcFace [11] & 78.7 & 84.3 & 77.2 & 75.1 & 83.4 & 86.1 & 76.9 & 83.8 & 81.6 \\ 
    \hline
    
    \end{tabular}
}
\end{table}

%% file: tables_suppl/table_synth_real_corruptions.tex
\begin{table}[t]
\caption{Synthetic to Real Benchmark with \rw augmentations\label{tab:synth2real_corr}}
\resizebox{1.0\textwidth}{!}{
    \begin{tabular}{l@{~}c@{~~}c|c@{~~}c|a@{~~}a|}
    \hline
    \multicolumn{7}{c|}{Synthetic to Real Benchmark - DGCNN~\cite{dgcnn}}\\
    \hline
    & \multicolumn{2}{c|}{SR 1} & \multicolumn{2}{c|}{SR 2} & \multicolumn{2}{a|}{Avg} \\
    \emph{Method} & {\scriptsize{AUROC}$\uparrow$} & {\scriptsize{FPR95}$\downarrow$} & {\scriptsize{AUROC}$\uparrow$} & {\scriptsize{FPR95}$\downarrow$} & {\scriptsize{AUROC}$\uparrow$} & {\scriptsize{FPR95}$\downarrow$} \\ \hline
      
    MSP~[18] & 72.2 & {91.0} & 61.2 & {90.3} & 66.7 & 90.6 \\
    MSP (+RW Augm)        & 82.1 & {76.0} & 65.8 & {92.8} & 73.9 & 84.4 \\ \hline
    SubArcFace~[11] & 74.5 & {86.7} & 68.7 &{86.6} & 71.6 & 86.7 \\
    SubArcFace (+RW Augm) & 81.3 & {77.4} & 68.8 & {84.7} & \textbf{75.1} & \textbf{81.1} \\ \hline
    \end{tabular}
    \begin{tabular}{|c@{~~}c|c@{~~}c|a@{~~}a}
    \hline
    \multicolumn{6}{|c}{Synthetic to Real Benchmark - PointNet++~\cite{pointnet++}}         \\ \hline
    \multicolumn{2}{|c|}{SR 1} & \multicolumn{2}{c|}{SR 2}  & \multicolumn{2}{a}{Avg} \\
    {\scriptsize{AUROC}$\uparrow$} & {\scriptsize{FPR95}$\downarrow$} & {\scriptsize{AUROC}$\uparrow$} & {\scriptsize{FPR95}$\downarrow$} & {\scriptsize{AUROC}$\uparrow$} & {\scriptsize{FPR95}$\downarrow$} \\ \hline
    81.0 & {79.6} & 70.3 & {86.7} & 75.6 & \textbf{83.2} \\
    76.5 & {81.8} & 74.6 & {85.9} & 75.5 & 83.9 \\ \hline
    78.7 & {84.3} & 75.1 & {83.4} & \textbf{76.9} & 83.8 \\
    76.9 & {83.7} & 73.0 & {89.5} & 75.0 & 86.6 \\ \hline
    \end{tabular}
}
\end{table}

%% file: tables_suppl/table_test_time_n_points.tex
\begin{table}[t]
\caption{Results on the Synthetic to Real Benchmark track when varying the number of test points. Reported results are average over the two possible scenarios (SR1, SR2).\label{tab:test_n_points}}
\resizebox{1.0\textwidth}{!}{
    \begin{tabular}{laa|aa||aa|aa||aa|aa}
    \hline
    & \multicolumn{4}{c||}{Synth (1024) to Real (2048) - Avg} & \multicolumn{4}{c||}{Synth (1024) to Real (1024) - Avg} & \multicolumn{4}{c}{Synth (1024) to Real (512) - Avg} \\
    \hline
    & \multicolumn{2}{a|}{DGCNN} & \multicolumn{2}{a||}{PointNet++} & \multicolumn{2}{a|}{DGCNN} & \multicolumn{2}{a||}{PointNet++} & \multicolumn{2}{a|}{DGCNN} & \multicolumn{2}{a}{PointNet++} \\
    \emph{Method} & {\scriptsize{AUROC}$\uparrow$} & {\scriptsize{FPR95}$\downarrow$} & {\scriptsize{AUROC}$\uparrow$} & {\scriptsize{FPR95}$\downarrow$} & {\scriptsize{AUROC}$\uparrow$} & {\scriptsize{FPR95}$\downarrow$} & {\scriptsize{AUROC}$\uparrow$} & {\scriptsize{FPR95}$\downarrow$} & {\scriptsize{AUROC}$\uparrow$} & {\scriptsize{FPR95}$\downarrow$} & {\scriptsize{AUROC}$\uparrow$} & {\scriptsize{FPR95}$\downarrow$}\\ \hline
    MSP [18]          & 66.7 & 90.6 & 75.6 & 83.2 & 70.2 & 86.7 & 76.5 & 84.0 & 58.5 & 92.1 & 74.3 & 84.4 \\
    MLS [44]          & 65.7 & 90.5 & 74.8 & 81.7 & 70.4 & 86.4 & 76.4 & 79.0 & 61.9 & 89.7 & 75.2 & 80.9 \\
    ODIN [27]         & 65.7 & 90.6 & 76.0 & 80.8 & 70.1 & 87.4 & 78.3 & 81.0 & 57.4 & 93.3 & 76.1 & 81.3 \\
    Energy [28]       & 65.6 & 90.8 & 74.8 & 82.4 & 70.6 & 86.9 & 77.6 & 78.9 & 62.5 & 88.7 & 76.3 & 80.4 \\
    GradNorm [21]     & 63.4 & 91.5 & 73.0 & 83.2 & 70.5 & 85.8 & 76.5 & 78.0 & 62.2 & 89.8 & 75.5 & 79.8 \\
    ReAct [40]        & 65.6 & 90.5 & 74.6 & 81.4 & 67.5 & 88.3 & 74.3 & 80.9 & 60.8 & 90.4 & 73.1 & 82.8 \\
    \hline
    Cosine proto & 57.9 & 90.9 & \textbf{78.2} & \textbf{76.1} & 70.0 & 86.3 & 76.2 & 77.7 & 61.1 & 88.8 & 74.9 & 79.5 \\
    \eucli      & 66.0 & 89.2 & 77.7 & 82.3 & \textbf{73.9} & \textbf{82.2} & 77.0 & 84.2 & \textbf{64.9} & \textbf{88.4} & 72.4 & 86.5 \\
    SubArcface [11]   & \textbf{71.6} & \textbf{86.7} & 76.9 & 83.8 & 61.9 & 88.6 & \textbf{78.5} & \textbf{76.4} & 55.0 & 93.1 & \textbf{76.8} & \textbf{76.6} \\
    \hline
    \end{tabular}
}
\end{table}

%% file: sections_suppl/supp-error_bars.tex
\section{Analysis of the error margin}
\begin{figure*}[t!]
\centering
\includegraphics[width=1.02\textwidth]{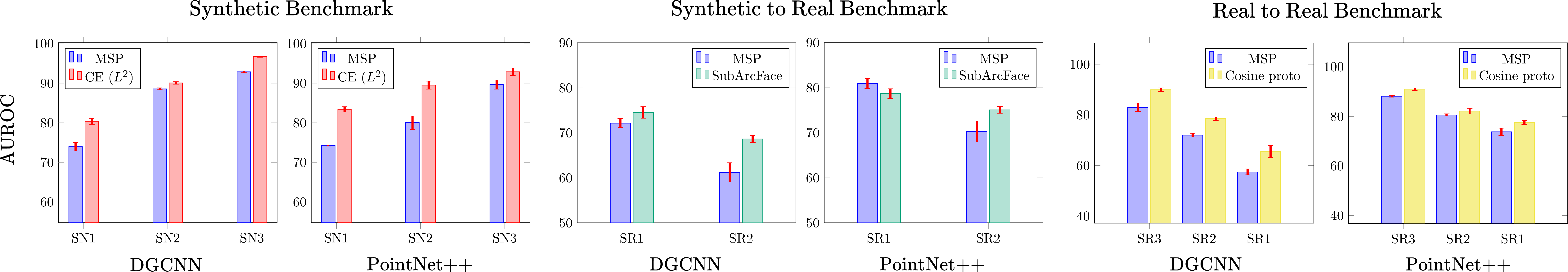}
\caption{Error margin analysis. State-of-the-art and baseline AUROC results on the three 3DOS tracks: Synthetic, Synthetic to Real and Real to Real Benchmarks. Red error bars represent the standard deviation around the mean value.\label{fig:error_bars}}
\end{figure*}

All the experimental results presented in the main paper are average over three experiments repetitions with different seeds.
In Fig. \ref{fig:error_bars} we report both the average and standard deviation for the MSP baseline and the methods that presented top results, respectively \eucli for the \emph{Synthetic}, SubArcFace for the \emph{Synthetic to Real} and Cosine proto for the \emph{Real to Real} Benchmarks.
By looking at the error bars, we can see that in the Synthetic Benchmark the standard deviation is quite small and the advantage of \eucli over MSP is always statistically significant. 
In the Synthetic to Real Benchmark the performance gap is lower, especially for the SR1 case, where MSP outperforms SubArcFace with the PointNet++ backbone. 
With both backbones however the results of the baseline and the state-of-the-art approach are within the error margin. 
SubArcface however has significantly better performance on the SR2 setting with also a smaller standard deviation.

When inspecting the Real to Real Benchmark results the performance gap between the baseline and the top-performing method is most noticeable with DGCNN.
Clearly, the improved performance obtained by using a more robust backbone such as PointNet++ reduces the importance of selecting a stronger learning approach.
In any case, the error margin is quite low and appears to decrease as performance improves, highlighting the high reliability of the results obtained in this Benchmark. 

%% file: sections_suppl/supp-limit.tex
\section{Further discussion}
\subsection{Limitations}
Some limitations of our work are directly inherited from the 3D computer vision field. 
The benchmark would undoubtedly benefit from the inclusion of a large-scale \rw dataset; however, such a dataset would have to be purposefully collected and curated because it is currently unavailable.
In the last years, huge progress has been made in 3D deep learning literature. However, most of the recent works exclusively focus on synthetic scenarios, now exhibiting a trend of performance saturation on these testbeds.
Furthermore, the lack of a large-scale annotated dataset also limits the development of more efficient 3D backbones.
Our experiments demonstrated that using a cutting-edge backbone does not automatically translate into improved performance.
This surprising trend is even more visible in the synthetic to \rw scenario, for which more research efforts are needed. 

\subsection{Broader Impact}
We hope that our research will have a positive impact on both academia and society.
In terms of academic research, we emphasized the importance of investigating Open Set scenarios for 3D deep learning.
In this context, our benchmark will serve as a reliable starting point for novel methods capable of leveraging the plethora of information naturally offered by 3D data.
We release our code with the aim of providing a foothold for future work towards building trustworthy systems that can manage the challenges of the open world. 
In terms of societal impact, we anticipate that increased academic interest in this field will drive the development of more robust models for safety-critical applications where 3D sensing could be a valuable ally: 
autonomous driving, robotics and health care are only some examples.

%% file: neurips_data_2022.bbl
\begin{thebibliography}{10}

\bibitem{abati2019latent}
D.~Abati, A.~Porrello, S.~Calderara, and R.~Cucchiara.
\newblock {Latent Space Autoregression for Novelty Detection}.
\newblock In {\em CVPR}, 2019.

\bibitem{alliegro_icpr}
A.~Alliegro, D.~Boscaini, and T.~Tommasi.
\newblock Joint supervised and self-supervised learning for 3d real world
  challenges.
\newblock In {\em ICPR}, 2021.

\bibitem{biplab1}
A.~Bhardwaj, S.~Pimpale, S.~Kumar, and B.~Banerjee.
\newblock Empowering knowledge distillation via open set recognition for robust
  3d point cloud classification.
\newblock {\em Pattern Recognition Letters}, 151:172--179, 2021.

\bibitem{cardace2021refrec}
A.~Cardace, R.~Spezialetti, P.~Z. Ramirez, S.~Salti, and L.~D. Stefano.
\newblock Refrec: Pseudo-labels refinement via shape reconstruction for
  unsupervised 3d domain adaptation.
\newblock In {\em 3DV}, 2021.

\bibitem{3dv-opensetdetection}
J.~Cen, P.~Yun, J.~Cai, M.~Wang, and M.~Liu.
\newblock Open-set 3d object detection.
\newblock In {\em 3DV}, 2021.

\bibitem{shapenet_dataset}
A.~X. Chang, T.~Funkhouser, L.~Guibas, P.~Hanrahan, Q.~Huang, Z.~Li,
  S.~Savarese, M.~Savva, S.~Song, H.~Su, J.~Xiao, L.~Yi, and F.~Yu.
\newblock {ShapeNet: An Information-Rich 3D Model Repository}.
\newblock {\em preprint arXiv:1512.03012}, 2015.

\bibitem{ARPL_TPAMI}
G.~Chen, P.~Peng, X.~Wang, and Y.~Tian.
\newblock Adversarial reciprocal points learning for open set recognition.
\newblock {\em IEEE Transactions on Pattern Analysis and Machine Intelligence},
  2021.

\bibitem{DRP_eccv2020}
G.~Chen, L.~Qiao, Y.~Shi, P.~Peng, J.~Li, T.~Huang, S.~Pu, and Y.~Tian.
\newblock Learning open set network with discriminative reciprocal points.
\newblock In {\em ECCV}, 2020.

\bibitem{chen2021atom}
J.~Chen, Y.~Li, X.~Wu, Y.~Liang, and S.~Jha.
\newblock Atom: Robustifying out-of-distribution detection using outlier
  mining.
\newblock In {\em ECML}, 2021.

\bibitem{Choi2020Novelty}
S.~Choi and S.-Y. Chung.
\newblock Novelty detection via blurring.
\newblock In {\em ICLR}, 2020.

\bibitem{subarcface}
J.~Deng, J.~Guo, T.~Liu, M.~Gong, and S.~Zafeiriou.
\newblock Sub-center arcface: Boosting face recognition by large-scale noisy
  web faces.
\newblock In {\em ECCV}, 2020.

\bibitem{arcface}
J.~Deng, J.~Guo, N.~Xue, and S.~Zafeiriou.
\newblock Arcface: Additive angular margin loss for deep face recognition.
\newblock In {\em CVPR}, 2019.

\bibitem{realnvp}
L.~Dinh, J.~Sohl-Dickstein, and S.~Bengio.
\newblock {Density estimation using Real NVP}.
\newblock In {\em ICLR}, 2017.

\bibitem{du2022vos}
X.~Du, Z.~Wang, M.~Cai, and Y.~Li.
\newblock {VOS: Learning What You Don’t Know by Virtual Outlier Synthesis}.
\newblock In {\em ICLR}, 2022.

\bibitem{fontanel2021detecting}
D.~Fontanel, F.~Cermelli, M.~Mancini, and B.~Caputo.
\newblock Detecting anomalies in semantic segmentation with prototypes.
\newblock In {\em CVPR-W}, 2021.

\bibitem{calibration_deep_nets}
C.~Guo, G.~Pleiss, Y.~Sun, and K.~Q. Weinberger.
\newblock On calibration of modern neural networks.
\newblock In {\em ICML}, 2017.

\bibitem{pct}
M.-H. Guo, J.-X. Cai, Z.-N. Liu, T.-J. Mu, R.~R. Martin, and S.-M. Hu.
\newblock Pct: Point cloud transformer.
\newblock {\em Computational Visual Media}, 7(2):187--199, 2021.

\bibitem{MSP}
D.~Hendrycks and K.~Gimpel.
\newblock A baseline for detecting misclassified and out-of-distribution
  examples in neural networks.
\newblock {\em ICLR}, 2017.

\bibitem{outlier_exposure}
D.~Hendrycks, M.~Mazeika, and T.~Dietterich.
\newblock Deep anomaly detection with outlier exposure.
\newblock In {\em ICLR}, 2019.

\bibitem{huang2021Euclideanfeature}
H.~Huang, Z.~Li, L.~Wang, S.~Chen, B.~Dong, and X.~Zhou.
\newblock Feature space singularity for out-of-distribution detection.
\newblock In {\em SafeAI-W}, 2021.

\bibitem{gradnorm}
R.~Huang, A.~Geng, and Y.~Li.
\newblock On the importance of gradients for detecting distributional shifts in
  the wild.
\newblock In {\em NeurIPS}, 2021.

\bibitem{khosla2020supervised}
P.~Khosla, P.~Teterwak, C.~Wang, A.~Sarna, Y.~Tian, P.~Isola, A.~Maschinot,
  C.~Liu, and D.~Krishnan.
\newblock Supervised contrastive learning.
\newblock In {\em NeurIPS}, 2020.

\bibitem{Kong_2021_ICCV}
S.~Kong and D.~Ramanan.
\newblock {OpenGAN: Open-Set Recognition via Open Data Generation}.
\newblock In {\em ICCV}, 2021.

\bibitem{Lee_2021_CVPR}
D.~Lee, J.~Lee, J.~Lee, H.~Lee, M.~Lee, S.~Woo, and S.~Lee.
\newblock Regularization strategy for point cloud via rigidly mixed sample.
\newblock In {\em CVPR}, 2021.

\bibitem{NEURIPS2018_abdeb6f5}
K.~Lee, K.~Lee, H.~Lee, and J.~Shin.
\newblock A simple unified framework for detecting out-of-distribution samples
  and adversarial attacks.
\newblock In {\em NeurIPS}, 2018.

\bibitem{Li_2020_CVPR}
Y.~Li and N.~Vasconcelos.
\newblock Background data resampling for outlier-aware classification.
\newblock In {\em CVPR}, 2020.

\bibitem{ODIN}
S.~Liang, Y.~Li, and R.~Srikant.
\newblock Enhancing the reliability of out-of-distribution image detection in
  neural networks.
\newblock In {\em ICLR}, 2018.

\bibitem{energy}
W.~Liu, X.~Wang, J.~Owens, and Y.~Li.
\newblock Energy-based out-of-distribution detection.
\newblock In {\em NeurIPS}, 2020.

\bibitem{RSCNN}
Y.~Liu, B.~Fan, S.~Xiang, and C.~Pan.
\newblock Relation-shape convolutional neural network for point cloud analysis.
\newblock In {\em CVPR}, 2019.

\bibitem{pointMLP}
X.~Ma, C.~Qin, H.~You, H.~Ran, and Y.~Fu.
\newblock Rethinking network design and local geometry in point cloud: A simple
  residual {MLP} framework.
\newblock In {\em ICLR}, 2022.

\bibitem{towards_japan}
M.~Masuda, R.~Hachiuma, R.~Fujii, H.~Saito, and Y.~Sekikawa.
\newblock Toward unsupervised 3d point cloud anomaly detection using
  variational autoencoder.
\newblock In {\em ICIP}, 2021.

\bibitem{counterfactual}
L.~Neal, M.~Olson, X.~Fern, W.-K. Wong, and F.~Li.
\newblock Open set learning with counterfactual images.
\newblock In {\em ECCV}, 2018.

\bibitem{nguyen2015deep}
A.~Nguyen, J.~Yosinski, and J.~Clune.
\newblock Deep neural networks are easily fooled: High confidence predictions
  for unrecognizable images.
\newblock In {\em CVPR}, 2015.

\bibitem{pointnet++}
C.~R. Qi, L.~Yi, H.~Su, and L.~J. Guibas.
\newblock Pointnet++: Deep hierarchical feature learning on point sets in a
  metric space.
\newblock In {\em NeurIPS}, 2017.

\bibitem{NEURIPS2019_ratios}
J.~Ren, P.~J. Liu, E.~Fertig, J.~Snoek, R.~Poplin, M.~Depristo, J.~Dillon, and
  B.~Lakshminarayanan.
\newblock Likelihood ratios for out-of-distribution detection.
\newblock In {\em NeurIPS}, 2019.

\bibitem{ruan2022optimal}
Y.~Ruan, Y.~Dubois, and C.~J. Maddison.
\newblock Optimal representations for covariate shift.
\newblock In {\em ICLR}, 2022.

\bibitem{RudWan2021differnet}
M.~Rudolph, B.~Wandt, and B.~Rosenhahn.
\newblock Same same but differnet: Semi-supervised defect detection with
  normalizing flows.
\newblock In {\em WACV}, 2021.

\bibitem{gramICML2020}
C.~S. Sastry and S.~Oore.
\newblock Detecting out-of-distribution examples with {G}ram matrices.
\newblock In {\em ICML}, 2020.

\bibitem{sehwag2021_SSD}
V.~Sehwag, M.~Chiang, and P.~Mittal.
\newblock {SSD: A Unified Framework for Self-Supervised Outlier Detection}.
\newblock In {\em ICLR}, 2021.

\bibitem{podn_scientificreport}
Y.~Shu, Y.~Shi, Y.~Wang, T.~Huang, and Y.~Tian.
\newblock {P-ODN}: Prototype-based open deep network for open set recognition.
\newblock {\em Nature Scientific Reports}, 10(1):1--13, 2020.

\bibitem{corruptions}
J.~Sun, Q.~Zhang, B.~Kailkhura, Z.~Yu, C.~Xiao, and Z.~M. Mao.
\newblock Benchmarking robustness of 3d point cloud recognition against common
  corruptions.
\newblock {\em preprint arXiv:2201.12296}, 2022.

\bibitem{react}
Y.~Sun, C.~Guo, and Y.~Li.
\newblock {ReAct: Out-of-distribution Detection With Rectified Activations}.
\newblock In {\em NeurIPS}, 2021.

\bibitem{LS}
C.~Szegedy, V.~Vanhoucke, S.~Ioffe, J.~Shlens, and Z.~Wojna.
\newblock Rethinking the inception architecture for computer vision.
\newblock In {\em CVPR}, 2016.

\bibitem{tack2020_CSI}
J.~Tack, S.~Mo, J.~Jeong, and J.~Shin.
\newblock {CSI: Novelty Detection via Contrastive Learning on Distributionally
  Shifted Instances}.
\newblock In {\em NeurIPS}, 2020.

\bibitem{scanobjectnn_ICCV19}
M.~A. Uy, Q.-H. Pham, B.-S. Hua, D.~T. Nguyen, and S.-K. Yeung.
\newblock {Revisiting Point Cloud Classification: A New Benchmark Dataset and
  Classification Model on Real-World Data}.
\newblock In {\em ICCV}, 2019.

\bibitem{vaze2022openset}
S.~Vaze, K.~Han, A.~Vedaldi, and A.~Zisserman.
\newblock {Open-Set Recognition: A Good Closed-Set Classifier is All You Need}.
\newblock In {\em ICLR}, 2022.

\bibitem{cosface}
H.~Wang, Y.~Wang, Z.~Zhou, X.~Ji, D.~Gong, J.~Zhou, Z.~Li, and W.~Liu.
\newblock Cosface: Large margin cosine loss for deep face recognition.
\newblock In {\em CVPR}, 2018.

\bibitem{dgcnn}
Y.~Wang, Y.~Sun, Z.~Liu, S.~E. Sarma, M.~M. Bronstein, and J.~M. Solomon.
\newblock {Dynamic Graph CNN for Learning on Point Clouds}.
\newblock {\em ACM Transactions on Graphics (TOG)}, 38(5):1--12, 2019.

\bibitem{WongWRLU19Urtasun}
K.~Wong, S.~Wang, M.~Ren, M.~Liang, and R.~Urtasun.
\newblock Identifying unknown instances for autonomous driving.
\newblock In {\em CoRL}, 2019.

\bibitem{mn40_dataset}
Z.~Wu, S.~Song, A.~Khosla, F.~Yu, L.~Zhang, X.~Tang, and J.~Xiao.
\newblock {3D ShapeNets: A Deep Representation for Volumetric Shapes}.
\newblock In {\em CVPR}, 2015.

\bibitem{curvenet}
T.~Xiang, C.~Zhang, Y.~Song, J.~Yu, and W.~Cai.
\newblock Walk in the cloud: Learning curves for point clouds shape analysis.
\newblock In {\em ICCV}, 2021.

\bibitem{NEURIPS2020_eddea82a}
Z.~Xiao, Q.~Yan, and Y.~Amit.
\newblock Likelihood regret: An out-of-distribution detection score for
  variational auto-encoder.
\newblock In {\em NeurIPS}, 2020.

\bibitem{GPointNet}
J.~Xie, Y.~Xu, Z.~Zheng, R.~Gao, W.~Wang, Z.~Song-Chun, and Y.~N. Wu.
\newblock {Generative PointNet: Deep Energy-Based Learning on Unordered Point
  Sets for 3D Generation, Reconstruction and Classification}.
\newblock In {\em CVPR}, 2021.

\bibitem{gdanet}
M.~Xu, J.~Zhang, Z.~Zhou, M.~Xu, X.~Qi, and Y.~Qiao.
\newblock Learning geometry-disentangled representation for complementary
  understanding of 3d object point cloud.
\newblock In {\em AAAI}, 2021.

\bibitem{yang2021scood}
J.~Yang, H.~Wang, L.~Feng, X.~Yan, H.~Zheng, W.~Zhang, and Z.~Liu.
\newblock Semantically coherent out-of-distribution detection.
\newblock In {\em ICCV}, 2021.

\bibitem{yang2021oodsurvey}
J.~Yang, K.~Zhou, Y.~Li, and Z.~Liu.
\newblock Generalized out-of-distribution detection: A survey.
\newblock {\em preprint arXiv:2110.11334}, 2021.

\bibitem{foldingnet}
Y.~Yang, C.~Feng, Y.~Shen, and D.~Tian.
\newblock {FoldingNet: Point Cloud Auto-Encoder via Deep Grid Deformation}.
\newblock In {\em CVPR}, 2018.

\bibitem{MCD}
Q.~Yu and K.~Aizawa.
\newblock Unsupervised out-of-distribution detection by maximum classifier
  discrepancy.
\newblock In {\em ICCV}, 2019.

\bibitem{OpenHybrid}
H.~Zhang, A.~Li, J.~Guo, and Y.~Guo.
\newblock Hybrid models for open set recognition.
\newblock In {\em ECCV}, 2020.

\bibitem{OE_mixup}
J.~Zhang, N.~Inkawhich, Y.~Chen, and H.~Li.
\newblock Fine-grained out-of-distribution detection with mixup outlier
  exposure.
\newblock {\em preprint arXiv:2106.03917}, 2021.

\end{thebibliography}
